\pdfoutput=1
\documentclass{article}
\usepackage{simauthor_arxiv,times}

\usepackage[T1]{fontenc}
\usepackage[utf8]{inputenc}
\usepackage{microtype}
\usepackage{amsmath,amssymb,mathtools}
\usepackage{booktabs,array,tabularx,multirow}
\usepackage{graphicx}
\usepackage[table]{xcolor}
\usepackage{xspace}
\usepackage{enumitem}
\usepackage{caption}
\usepackage{float}
\usepackage{listings}
\usepackage{hyperref}
\usepackage{pifont}
\usepackage{algorithm}
\usepackage{algpseudocode}
\usepackage{tcolorbox}
\definecolor{oodgreen}{RGB}{241,248,241}
\usepackage{array}

\usepackage{fix-cm}
\usepackage{xcolor}
\usepackage{booktabs}
\usepackage{graphicx}
\usepackage{tikz}
\usepackage{listings}
\usepackage{tcolorbox}
\tcbuselibrary{skins,listings}

\definecolor{saPanelBg}{RGB}{249,249,249}
\definecolor{saMuted}{RGB}{108,113,121}
\definecolor{saDiffRed}{RGB}{166,55,55}
\definecolor{saDiffRedBg}{RGB}{252,241,241}
\definecolor{saDiffGreen}{RGB}{36,116,70}
\definecolor{saDiffGreenBg}{RGB}{238,248,241}
\definecolor{saCodeKeyword}{RGB}{25,55,105}
\definecolor{saCodeString}{RGB}{95,52,38}

\newcommand{\saPanelH}{1.25in}

\lstdefinestyle{saDiffCode}{
  language=Python,
  basicstyle=\ttfamily\fontsize{5.0pt}{5.6pt}\selectfont,
  keywordstyle=\bfseries\color{saCodeKeyword},
  commentstyle=\itshape\color{saMuted},
  stringstyle=\color{saCodeString},
  showstringspaces=false,
  columns=fullflexible,
  keepspaces=true,
  tabsize=2,
  breaklines=false,
  numbers=none,
  aboveskip=0pt,
  belowskip=0pt
}

\newtcblisting{saBeforeBlock}{
  enhanced,
  listing only,
  listing engine=listings,
  listing options={style=saDiffCode},
  colback=saDiffRedBg,
  colframe=saDiffRedBg,
  boxrule=0pt,
  arc=1.2pt,
  outer arc=1.2pt,
  left=2.5pt,
  right=2pt,
  top=2pt,
  bottom=2pt,
  boxsep=0pt,
  before skip=0pt,
  after skip=0pt
}

\newtcblisting{saAfterBlock}{
  enhanced,
  listing only,
  listing engine=listings,
  listing options={style=saDiffCode},
  colback=saDiffGreenBg,
  colframe=saDiffGreenBg,
  boxrule=0pt,
  arc=1.2pt,
  outer arc=1.2pt,
  left=2.5pt,
  right=2pt,
  top=2pt,
  bottom=2pt,
  boxsep=0pt,
  before skip=0pt,
  after skip=0pt
}

\newcommand{\saDiffHead}[2]{%
  {\fontsize{5.3pt}{5.8pt}\selectfont\bfseries\color{#1}#2}%
  \par\vspace{1.1pt}%
}

\newtcolorbox{saCodePanel}[1]{
  enhanced,
  width=\linewidth,
  height=\saPanelH,
  colback=saPanelBg,
  colframe=black,
  coltitle=white,
  colbacktitle=black,
  title={#1},
  fonttitle=\bfseries\fontsize{6.5pt}{7.1pt}\selectfont,
  boxrule=0.75pt,
  arc=2pt,
  outer arc=2pt,
  left=3.5pt,
  right=3.5pt,
  top=3.0pt,
  bottom=3.0pt,
  boxsep=0pt,
  titlerule=0pt,
  toptitle=2.2pt,
  bottomtitle=2.2pt,
  before skip=0pt,
  after skip=0pt
}

\newtcolorbox{saTablePanel}[1]{
  enhanced,
  width=\linewidth,
  height=\saPanelH,
  valign=center,
  colback=saPanelBg,
  colframe=black,
  coltitle=white,
  colbacktitle=black,
  title={#1},
  fonttitle=\bfseries\fontsize{6.5pt}{7.1pt}\selectfont,
  boxrule=0.75pt,
  arc=2pt,
  outer arc=2pt,
  left=3.0pt,
  right=3.0pt,
  top=2.6pt,
  bottom=2.6pt,
  boxsep=0pt,
  titlerule=0pt,
  toptitle=2.2pt,
  bottomtitle=2.2pt,
  before skip=0pt,
  after skip=0pt
}

\definecolor{posred}{HTML}{C53030}   
\definecolor{neggreen}{HTML}{276749} 

\hypersetup{
  hidelinks,
  pdftitle={SimAuthor: Harnessing Foundation Models for Persistent Scientific Simulator Authoring},
  pdfauthor={Yishan Wang, Ran Piao, Mathias Funk, Aaqib Saeed}
}

\setlist{nosep,leftmargin=*}

\newcommand{\ra}[1]{\renewcommand{\arraystretch}{#1}}
\newcolumntype{R}{>{\raggedleft\arraybackslash}X}
\lstdefinestyle{diffstyle}{basicstyle=\ttfamily\scriptsize,breaklines=false,
  columns=fullflexible,frame=none,xleftmargin=8pt,
  aboveskip=3pt,belowskip=5pt,showstringspaces=false}

\newcommand{\framework}{\textsc{SimAuthor}\xspace}

\newcommand{\appref}[1]{Appendix~\ref{#1}}

\newcommand{\slot}[1]{\texttt{\{#1\}}}

\title{\framework: Harnessing Foundation Models for Persistent Scientific Simulator Authoring\thanks{Code: \url{https://github.com/yishani/SimAuthor}. Project page: \url{https://yishani.github.io/SimAuthor}.}}
\author{%
Yishan Wang \qquad Ran Piao \qquad Mathias Funk \qquad Aaqib Saeed\\
Eindhoven University of Technology
}
\date{}

\begin{document}
\maketitle

\begin{abstract}
Foundation models can generate scientific code, but authoring a scientific simulator (an executable program encoding hypotheses about how mechanisms generate observable signals) requires iterative refinement. Scientific adequacy rarely admits a unique implementation or exact test, so simulators must instead
be judged against limited real observations.
We study this setting as \emph{scientific simulator authoring under weak empirical feedback}, where distributional comparisons between simulated and real signals guide revision, and the target is the simulator itself rather than only its generated samples. 
We introduce \framework{}, a persistent authoring harness that retains and revises executable simulators, separates scalar search scores from structured discrepancy feedback, and accumulates reusable implementation mechanisms. We evaluate \framework{} on six biomedical tasks spanning cardiac and respiratory audio, photoplethysmography (PPG), and electrocardiography (ECG). Under a fixed 100-attempt budget, \framework{} outperforms PUCT score search on all six tasks, generally outperforms textual-strategy optimization, and achieves the highest endpoint score on five of six. The authored simulators also improve on unseen recordings, transfer to independent pretrained representations, and yield
substantial out-of-distribution gains in downstream ECG classification. Finally, 111 of 138 audited revisions alter program structure and account for 86.1\% of the signed score improvement. These results suggest that persistent revision can progressively convert foundation-model knowledge into better executable
scientific simulators from limited empirical evidence.
\end{abstract}

\section{Introduction}

Foundation models, when coupled with inference-time harnesses that organize search, execution, and revision, are becoming increasingly capable scientific and programming systems. Pretraining equips them with substantial domain
knowledge and strong zero-shot code generation abilities
~\citep{singhal2023large,li2023starcoder,guo2024deepseek}, while inference-time scaffolding extends what they can accomplish through planning and iterative refinement
~\citep{shinn2023reflexion,romera2024mathematical,novikov2025alphaevolve}.
These capabilities raise an underexplored question: \emph{can foundation models author scientific simulators?} We consider a simulator an executable program that encodes hypotheses about how interpretable mechanisms and stochastic processes generate observable real-world signals. Unlike many LLM-guided
program-search settings, simulator authoring typically lacks an exact verifier: there is rarely a unique correct implementation, and no unit test determines whether the generated behavior is scientifically adequate. Candidate simulators must therefore be judged indirectly against limited real-world observations.

\begin{figure*}[t]
    \centering
    \includegraphics[width=\textwidth]{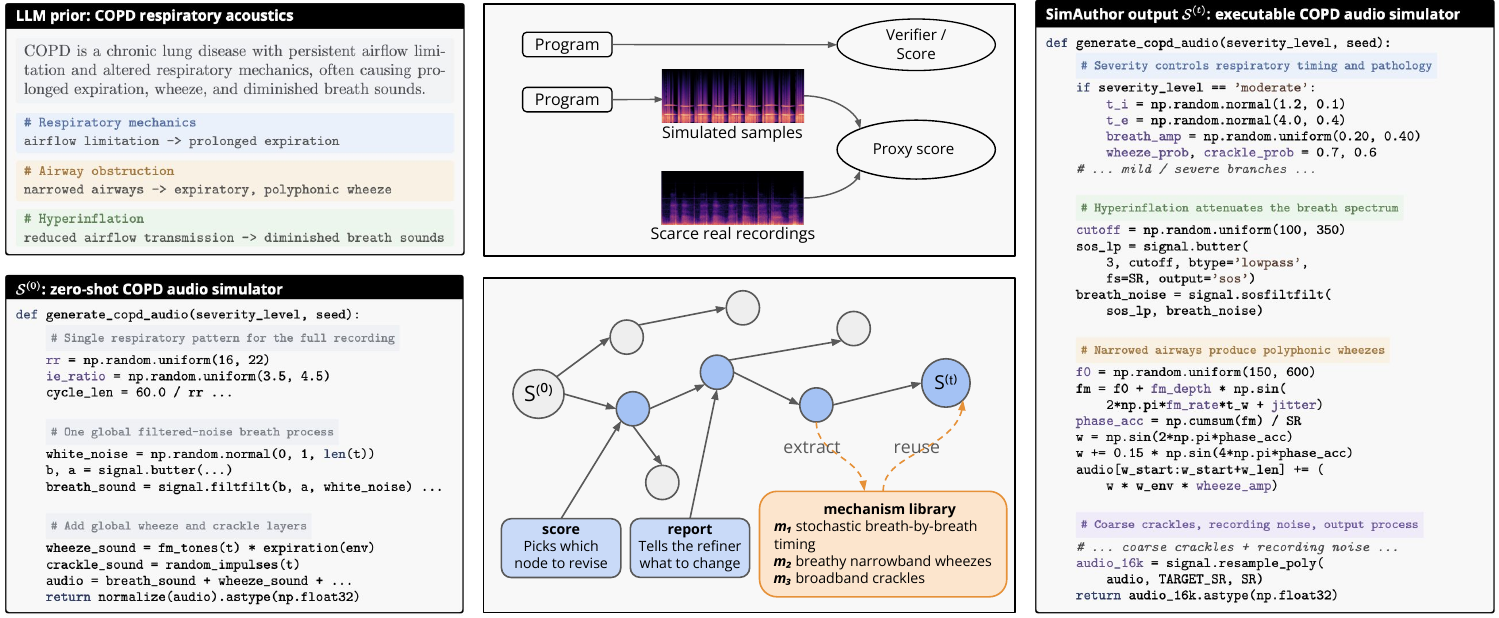}
    \caption{\textbf{Overview.} SimAuthor turns LLM prior knowledge into executable simulator code when exact verification is unavailable and feedback comes only from scarce real data.}
    \label{fig:splash}
\end{figure*}

We study this problem in \textit{biomedical signal simulation}, where the opportunity and the difficulty are especially concrete. For example, a COPD respiratory simulator is an executable program that generates breathing recordings from explicit models of respiratory timing, adventitious sounds, and measurement variability. Yet this knowledge does not directly translate into a useful simulator: a plausible zero-shot program may capture broad phenomenological features while omitting the stochastic, temporal, and measurement processes (Figure~\ref{fig:splash}, left). Simulation is particularly valuable when real recordings are scarce, imbalanced, or difficult to share~\citep{rieke2020federated,wagner2020ptbxl,rocha2017respiratory}, but the same scarcity provides only limited empirical grounding for authoring and validating the simulator. In the absence of an exact verifier, we use distributional comparisons between generated and real recordings as empirical feedback for revision (Figure~\ref{fig:splash}, middle top). In this work, the target is the simulator itself rather than only its samples: an explicit, executable model that can be inspected, modified, and controlled (Figure~\ref{fig:splash}, right).

We introduce \framework, a harness designed for this weak-feedback authoring regime (Figure~\ref{fig:splash}, middle). Rather than repeatedly generating programs from scratch, \framework persistently develops executable simulator
candidates, using empirical feedback to guide revision while retaining useful mechanisms discovered across model calls. In this way, simulator construction becomes a cumulative authoring process rather than zero-shot code generation.

We evaluate this setting on six biomedical simulator-authoring tasks spanning cardiac audio, respiratory audio, electrocardiography (ECG), and photoplethysmography (PPG). Under a fixed 100-attempt authoring budget, \framework{} outperforms repeated independent generation and score-only program
search on all six tasks, generally outperforms textual-strategy optimization, and achieves the strongest endpoint on five of six. Additional evaluations on unseen representations and downstream prediction tasks provide evidence that
these gains extend beyond the authoring objective. Component ablations show that structured discrepancy feedback is critical for final quality, while mechanism memory consistently improves search efficiency. A revision audit further finds that structural program changes account for 86.1\% of the signed
score improvement, indicating that progress is realized through changes to the explicit, inspectable simulator rather than only parameter retuning. These results support persistent artifact-centered revision as a way to convert foundation-model knowledge into better executable scientific models under weak
empirical feedback.

Our contributions can be summarized as follows:
\begin{enumerate}
    \item We introduce \textbf{foundation-model-based scientific simulator authoring} as a new problem setting: foundation models construct executable simulators that encode scientific assumptions and generative mechanisms. We formalize this setting under weak empirical feedback, where only limited real observations and no exact verifier are available.

    \item We introduce \framework, a persistent authoring harness that progressively revises executable simulators while carrying useful information and mechanisms across model calls.

    \item We evaluate simulator authoring across six biomedical tasks and four signal modalities, showing consistent gains over independent sampling, score-only search, and a textual-strategy optimization baseline, with additional evidence that these gains extend beyond the authoring objective.
\end{enumerate}

\section{Persistent Simulator Authoring}
\label{sec:method}

Let $\boldsymbol\theta\in\Theta$ denote the simulator's explicit variables, $\mathbf z\in\mathcal Z$ the values it samples at run time, such as beat-to-beat timing or recording noise, and $\mathbf y\in\mathcal Y$ an observable signal. A simulator is an executable program
\begin{equation}
\mathcal S:\Theta\times\mathcal Z\rightarrow\mathcal Y.
\end{equation}
The target of authoring is the program $\mathcal S$, not an individual sample. Authoring begins from a small reference set $\mathcal D_{\mathrm{ref}}$ of real recordings and a root simulator $\mathcal S^{(0)}$ written in a single call. Every subsequent attempt revises an existing candidate, so a trajectory has the form
\begin{equation}
\mathcal S^{(0)}\rightarrow\mathcal S^{(1)}\rightarrow\cdots\rightarrow\mathcal S^{(T)},
\end{equation}
where each revision can change program structure, parameters, which quantities are sampled and how, and how the signal is recorded (sensor response, filtering, noise). A single model call decides only what to write next. Everything else is the harness's to decide: which of the accumulated candidates to revise, what evidence about that candidate the model is shown, and what from earlier attempts is carried forward. These three decisions define the three components below.

\begin{figure}[ht]
    \centering
    \includegraphics[width=\linewidth]{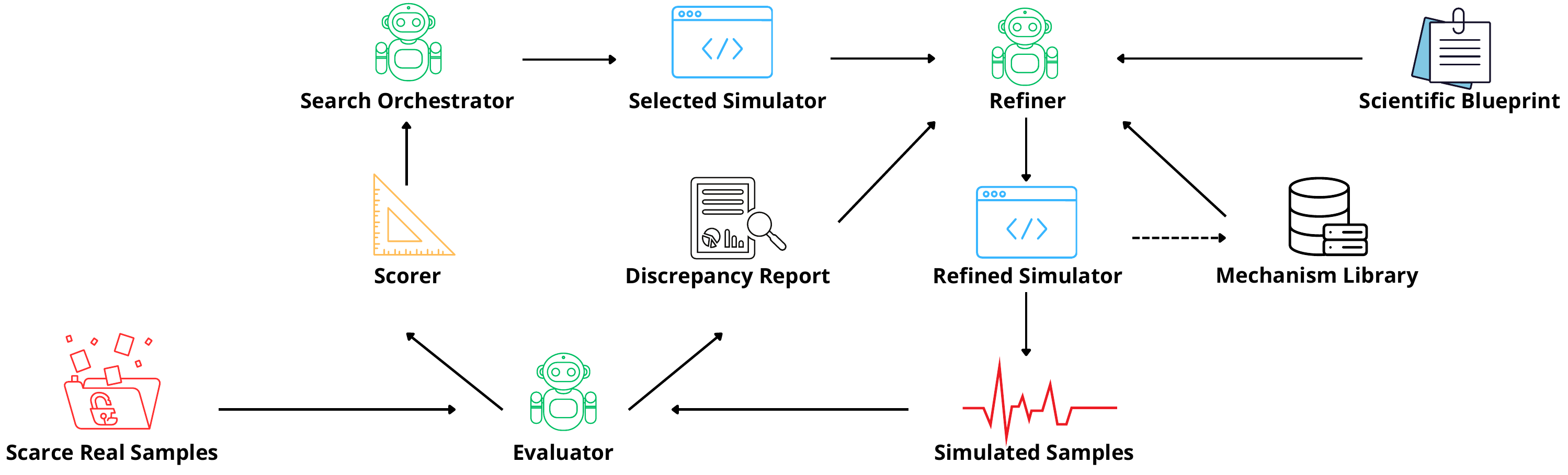}
    \caption{\textbf{The \framework authoring harness.} A search orchestrator selects simulator candidates based on numerical scores, while the refiner updates them using the scientific blueprint, discrepancy report, and mechanism library.
}
    \label{fig:framework}
\end{figure}

\subsection{Search score and discrepancy report}
\label{sec:feedback}

Each candidate is executed to generate a set of simulated observations, and a modality-specific \textbf{evaluator} compares these with the real reference observations, returning two complementary outputs. The \textbf{search score} is a scalar used by the search policy to select candidate programs. The \textbf{discrepancy report} is structured evidence describing where generated and real signals disagree. The distinction is deliberate: the score answers which candidate should search revise next, the report answers what should the refiner change.

\textbf{Evaluator.}
We fix one evaluator for each modality before
  search and never update it using generated programs. Audio and PPG are evaluated
  in signal space: the audio evaluator averages marginal distribution overlap for
  13 MFCCs and the zero-crossing rate, while the PPG evaluator compares ten
  pulse-level rhythm and morphology statistics. Because delineation-based
  objectives were brittle for ECG, we instead compare distributions of frozen
  ECGFounder embeddings using a scaled energy-distance score
  \citep{li2025ecgfounder}. All scores lie in $[0,1]$, with higher values indicating
  greater similarity to the reference set. The discrepancy report exposes the
  components underlying each score, such as feature-level mismatches, distribution
  spread, and possible mode collapse. These evaluators are fixed authoring
  objectives rather than complete measures of physiological fidelity; we therefore
  separately test generalization to held-out recordings, alternative
  representations, and downstream utility in Sections~\ref{sec:generalize}
  and~\ref{sec:heldout}.
  Full definitions are provided in Appendix~\ref{app:evaluator}.

\subsection{Scientific blueprint and mechanism library}
\label{sec:memory}

The refiner is conditioned on two complementary textual contexts: a \emph{static scientific blueprint} that anchors revision in domain knowledge, and a \emph{dynamic mechanism library} that accumulates useful implementation knowledge over the course of authoring. For each task, the scientific blueprint provides fixed context about relevant physiology, signal manifestations, and implementation constraints. It is generated by the foundation model before any recording is consulted, externalizing the model's prior knowledge as a document available to the
refiner on every attempt. The blueprint complements empirical discrepancy feedback: the latter indicates how simulated outputs differ from real recordings, while the former provides a stable scientific reference for how those discrepancies might be addressed. The prompt for blueprint generation is provided in \appref{app:blueprint}.

The mechanism library, in contrast, evolves during authoring. It stores reusable implementation mechanisms distilled from productive parent--child revisions. When a revision improves clearly over its parent, the program difference is summarized as a named mechanism, and recent mechanisms are made
available to subsequent refinement attempts. This allows implementation knowledge discovered in one branch of the search to persist and be reused later.
Representative mechanism-library entries, implementation details and full prompts for mechanism extraction are provided in
\appref{app:mechlib}, \appref{app:implementation}, and
\appref{app:extraction}.

\subsection{Search orchestration}
The orchestrator holds every candidate authored so far together with the tree of revisions that produced them. Following \citet{aygun2025ai}, we use a flat-pool PUCT search that balances exploitation of high-scoring programs with exploration of less-visited ones:
\begin{equation}
 n_t=\arg\max_{n\in\mathcal T}\left[
 \tilde v(n)+c_{\mathrm{puct}}\,p(n)
 \frac{\sqrt{N_{\mathrm{tot}}}}{1+N(n)}
 \right],
\end{equation}
where $\tilde v(n)\in[0,1]$ is the candidate's rank-normalized search score,
$N(n)$ its visit count, $N_{\mathrm{tot}}$ the total visits, and
$p(n)=1/|\mathcal T|$ a uniform prior. The first term favors high-scoring candidates and the second favors less-visited ones. PUCT score search uses the same selection rule, so comparing it with \framework{} holds search orchestration fixed and varies the information available to the refiner.
The end-to-end loop is given in \appref{app:algorithm}.

\section{Experimental Setup}
\label{sec:setup}
\paragraph{Tasks.} We evaluate six simulator-authoring tasks spanning four signal modalities: ventricular septal defect (VSD) and aortic stenosis (AS) in cardiac audio, chronic obstructive pulmonary disease (COPD) in respiratory audio, atrial fibrillation (AF) in PPG, and long-QT syndrome (LQT) and Wolff--Parkinson--White syndrome (WPW) in two-lead ECG. Several conditions are rare in public data: PTB-XL~\citep{wagner2020ptbxl} contains only 118 LQT and 80 WPW recordings among 21,837 ECGs. 
Each task has a small search-reference set (37--96 recordings), which the evaluator compares against during search, and a held-out evaluation set never seen during search: independent external datasets for four tasks, held-out same-source patient-disjoint splits for two. Details are in \appref{app:datasets}.

\paragraph{Baselines.}
We compare \framework{} against three alternatives spanning independent
sampling, score-guided program search, and natural-language optimization.
\textbf{Independent Sampling (Sampling)} spends the authoring budget on independently
generated programs and returns the best-of-$N$, following the repeated-sampling
strategy commonly used in code generation \citep{chen2021evaluating}.
\textbf{PUCT Score Search} selects parents with rank-based flat PUCT and revises each selected program using only its scalar score as feedback, following prior score-guided search over executable programs~\citep{aygun2025ai}.
\textbf{Textual Strategy Optimization (Text-Opt).}
Text-Opt uses the same blueprint, evaluator feedback, visual feedback, and PUCT policy as \framework{}, but searches over natural-language strategies rather than programs. Each node carries a strategy $\sigma_p$, from which the LM generates an updated strategy and a new simulator from scratch:
  \[
  (\sigma',P') \sim \operatorname{LM}(\sigma_p,B,O,R_p,V_p).
  \]
By contrast, \framework{} inherits and revises the selected program $P_p$ and can reuse mechanisms $M$ discovered across revisions:
  \[
  P' \sim \operatorname{LM}(P_p,M,B,O,R_p,V_p).
  \]
Thus, Text-Opt isolates strategy-space optimization from program inheritance and mechanism reuse \citep{yang2024opro,agrawal2025gepa}.

\paragraph{Authoring budget.} One authoring attempt is one foundation-model call proposing an executable simulator: for the search methods it revises an existing candidate, for the independent sampling baseline it is an independent program generation. Every method receives the same 100 attempts, and invalid or non-executable code consumes the budget. All primary trajectories use the same foundation model, Gemini 3.1 Pro~\citep{deepmind2026gemini31pro}. Numerical settings are in \appref{app:implementation}.


\begin{figure}[t]
\centering

\includegraphics[width=\linewidth]{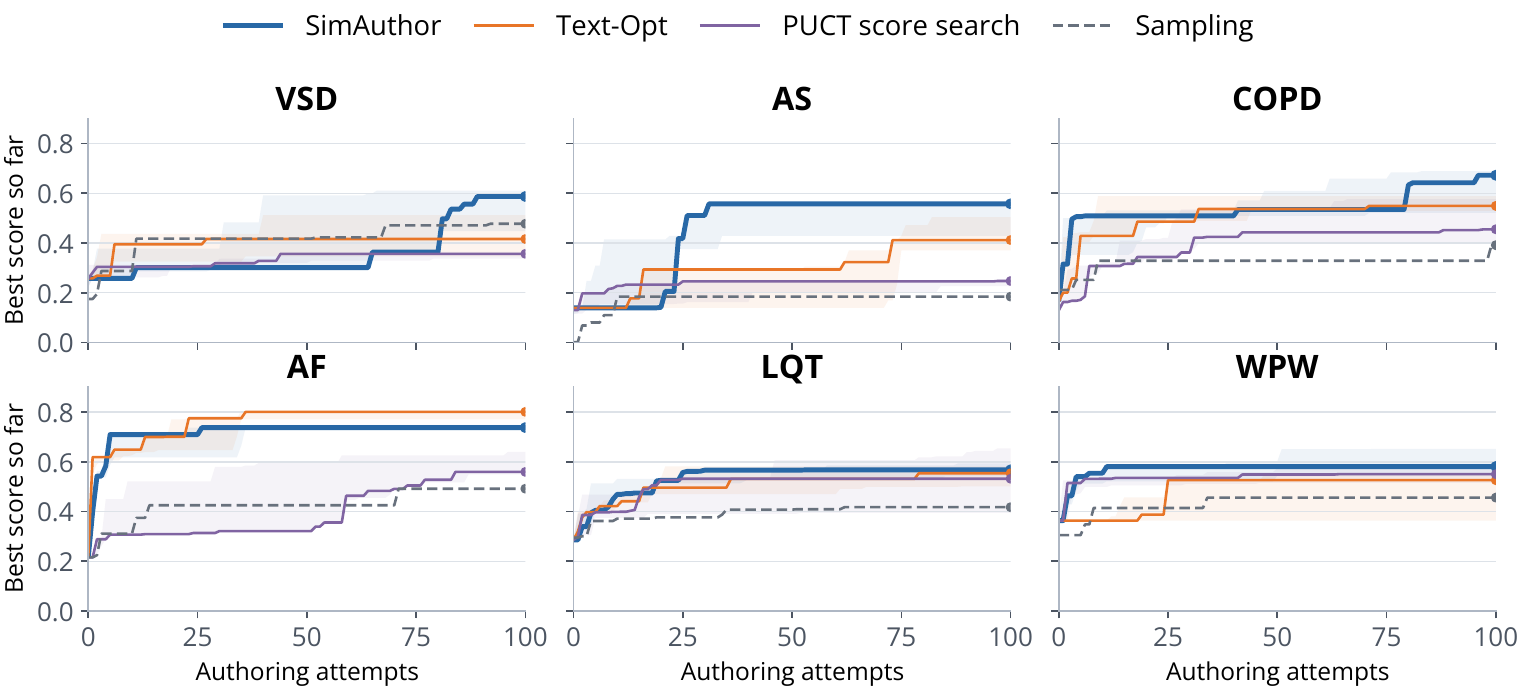}

\caption{\textbf{Best-so-far performance over 100 authoring attempts.}
  For seeded methods, curves show the complete trajectory of the median-Best@100
  run, with min--max shading across three seeds. Independent sampling follows its fixed
  proposal order.}
\label{fig:baseline}

\vspace{0.8\baselineskip}

\begin{minipage}{\linewidth}

\captionof{table}{
\textbf{Endpoint and upper-tail performance after 100 authoring attempts.}
Audio tasks use MFCC feature overlap, AF uses hand-designed PPG rhythm and
morphology statistics, and ECG tasks use frozen ECGFounder embeddings
\citep{li2025ecgfounder}.
Entries report Best@100 / top-10 mean. Seeded methods are summarized by the
median over three seeds; Sampling uses the fixed first 100 proposals.
Bold denotes the best result and underline the second best.
}
\label{tab:main}

\centering
\small
\setlength{\tabcolsep}{2.8pt}
\renewcommand{\arraystretch}{1.06}

\begin{tabular*}{0.98\textwidth}{
@{\extracolsep{\fill}}
l
cc
cc
cc
cc
cc
cc
@{}
}

\toprule

& \multicolumn{6}{c}{\textit{MFCC}}
& \multicolumn{2}{c}{\textit{Morphology}}
& \multicolumn{4}{c}{\textit{ECGF12}} \\
\cmidrule(lr){2-7}
\cmidrule(lr){8-9}
\cmidrule(lr){10-13}

\textbf{Method}
& \multicolumn{2}{c}{\textbf{VSD}}
& \multicolumn{2}{c}{\textbf{AS}}
& \multicolumn{2}{c}{\textbf{COPD}}
& \multicolumn{2}{c}{\textbf{AF}}
& \multicolumn{2}{c}{\textbf{LQT}}
& \multicolumn{2}{c}{\textbf{WPW}} \\[-1pt]

\addlinespace[1.5pt]

& \multicolumn{2}{c}{\scriptsize  $\mathcal{S}^{(0)}$: .258}
& \multicolumn{2}{c}{\scriptsize  $\mathcal{S}^{(0)}$: .139}
& \multicolumn{2}{c}{\scriptsize  $\mathcal{S}^{(0)}$: .165}
& \multicolumn{2}{c}{\scriptsize  $\mathcal{S}^{(0)}$: .216}
& \multicolumn{2}{c}{\scriptsize  $\mathcal{S}^{(0)}$: .286}
& \multicolumn{2}{c}{\scriptsize  $\mathcal{S}^{(0)}$: .364} \\[-1pt]

\cmidrule(lr){2-3}
\cmidrule(lr){4-5}
\cmidrule(lr){6-7}
\cmidrule(lr){8-9}
\cmidrule(lr){10-11}
\cmidrule(lr){12-13}

& \scriptsize $S_{\max}$ & \scriptsize $S_{\mathrm{top10}}$
& \scriptsize $S_{\max}$ & \scriptsize $S_{\mathrm{top10}}$
& \scriptsize $S_{\max}$ & \scriptsize $S_{\mathrm{top10}}$
& \scriptsize $S_{\max}$ & \scriptsize $S_{\mathrm{top10}}$
& \scriptsize $S_{\max}$ & \scriptsize $S_{\mathrm{top10}}$
& \scriptsize $S_{\max}$ & \scriptsize $S_{\mathrm{top10}}$ \\

\midrule

Sampling
& \underline{.477} & \underline{.424}
& .184 & .152
& .391 & .259
& .492 & .412
& .418 & .401
& .456 & .406 \\

PUCT Score Search
& .356 & .327
& .247 & .202
& .455 & .444
& .560 & .528
& .532 & \underline{.530}
& \underline{.550} & \textbf{.546} \\

Text-Opt
& .416 & .348
& \underline{.412} & \underline{.250}
& \underline{.549} & \underline{.505}
& \textbf{.801} & \textbf{.770}
& \underline{.554} & .477
& .526 & .457 \\

\addlinespace[1.5pt]

\textbf{SimAuthor}
& \textbf{.587} & \textbf{.483}
& \textbf{.557} & \textbf{.443}
& \textbf{.672} & \textbf{.628}
& \underline{.738} & \underline{.701}
& \textbf{.568} & \textbf{.553}
& \textbf{.581} & \underline{.527} \\
\bottomrule\end{tabular*}

\end{minipage}

\end{figure}

\section{Results}
\label{sec:results}

We organize the experiments around three questions.
First, does SimAuthor discover higher-quality simulators than independent sampling, score-guided program search, and textual strategy optimization, and how efficiently does it reach them (Section~\ref{sec:mainresult})? 
Second, do the resulting simulators generalize beyond the feedback used during authoring (Section~\ref{sec:generalize})?
Third, which parts of the authoring harness account for these gains (Section~\ref{sec:ablation})?

\subsection{SimAuthor pushes the simulator quality frontier}
\label{sec:mainresult}

We run each search-based method with three random seeds to capture run-to-run variation. To visualize one coherent trajectory, Figure~3 shows the run with the median final Best@100, with min--max shading across the three seeds. Table~1 reports $S_{\max}$ and $S_{\mathrm{top10}}$ from the same run; complete seed-level results are reported in Appendix~F.1.
\framework{} achieves the highest $S_{\max}$ on five of six tasks
(Figure~\ref{fig:baseline}, Table~\ref{tab:main}). AF is the exception: Text-Opt reaches
$.801$, compared with $.738$ for \framework{}. The top-10 results sharpen this picture. \framework{} has the highest $S_{\mathrm{top10}}$ on four tasks, showing that its advantage often extends beyond a single high-scoring endpoint
to the quality of the candidate pool; Text-Opt remains strongest on AF, while PUCT Score Search is slightly higher on WPW.

The controlled comparisons further indicate that search allocation alone does not explain these gains. Against PUCT Score Search, which uses the same candidate-selection rule, \framework{} improves $S_{\max}$ on all six tasks
($.031$--$.310$). Under weak empirical feedback, the limiting factor therefore appears to be not only which candidate
is revisited, but whether the refiner receives enough structured information for useful changes to accumulate in the retained program. Compared with Text-Opt, \framework{} is stronger on five of six tasks, suggesting that making the executable artifact itself persistent is generally more effective than carrying forward only a textual strategy and regenerating the program from scratch. AF is the lone exception. Its evaluator exposes physiologically  interpretable rhythm statistics whose discrepancies map relatively directly to synthesis changes, making textual strategy updates particularly effective.

Equal authoring attempts do not imply equal inference cost: \framework{} uses approximately $3.8\times$ the tokens of best-of-100 independent sampling because it carries substantially more context across revisions. Expanding independent sampling to the corresponding token-matched budget of roughly 380 proposals
does not close the gap, indicating that the gains are not explained by additional inference budget alone (\appref{app:compute}).

\subsection{SimAuthor's gains persist beyond the authoring objective}
\label{sec:generalize}

\begin{table}[t]
\centering
\caption{
\textbf{Generalization beyond the authoring objective.}
\textbf{(a)} Held-out representation evaluation on audio and PPG;
entries are $R_{\mathrm{sel}}\,|\,R_{\max}$.
\textbf{(b)} ECG downstream utility with 300-example augmentation,
reported as AUPRC; evaluation is in-distribution (ID) on PTB-XL and
out-of-distribution (OOD) on Chapman--Shaoxing.
Corresponding AUROC results in Appendix \ref{app:ecg-downstream}.
Bold indicates the best result.
}
\label{tab:generalization}

\footnotesize
\renewcommand{\arraystretch}{1.04}

\begin{minipage}[t]{0.58\textwidth}
\vspace{0pt}
\centering
\textbf{(a) Independent representation evaluation}\par\vspace{2pt}

\setlength{\tabcolsep}{0.8pt}

\begin{tabular*}{\linewidth}{
@{\extracolsep{\fill}}
lcccc
@{}
}
\toprule
& \multicolumn{3}{c}{\textit{OPERA-CE}}
& \textit{PaPaGei} \\
\cmidrule(lr){2-4}
\cmidrule(l){5-5}

\textbf{Method}
& \textbf{VSD}
& \textbf{AS}
& \textbf{COPD}
& \textbf{AF} \\
\midrule

Root $\mathcal{S}^{(0)}$
& $.317\,|\, .317$
& $.192\,|\, .192$
& $.152\,|\, .152$
& $.354\,|\, .354$ \\

Sampling
& $.123\,|\, .715$
& $.106\,|\, .612$
& $.162\,|\, .799$
& $.434\,|\, \mathbf{.830}$ \\

Text-Opt
& $\mathbf{.443}\,|\, .784$
& $.258\,|\, .685$
& $.468\,|\, \mathbf{.847}$
& $.631\,|\, .734$ \\

\textbf{SimAuthor}
& $.204\,|\, \mathbf{.789}$
& $\mathbf{.306}\,|\, \mathbf{.906}$
& $\mathbf{.561}\,|\, .752$
& $\mathbf{.665}\,|\, .737$ \\

\bottomrule
\end{tabular*}
\end{minipage}
\hfill
%
\begin{minipage}[t]{0.40\textwidth}
\vspace{0pt}
\centering
\textbf{(b) ECG downstream utility}\par\vspace{2pt}

\setlength{\tabcolsep}{1.5pt}

\begin{tabular*}{\linewidth}{
@{\extracolsep{\fill}}
l
c
>{\columncolor{oodgreen}}c
c
>{\columncolor{oodgreen}}c
@{}
}
\toprule
& \multicolumn{2}{c}{\textbf{LQT}}
& \multicolumn{2}{c}{\textbf{WPW}} \\
\cmidrule(lr){2-3}
\cmidrule(l){4-5}

\textbf{Training}
& \textbf{ID}
& \textbf{OOD}
& \textbf{ID}
& \textbf{OOD} \\
\midrule

Real
  & $\mathbf{.406}$
  & $.633$
  & $.857$
  & $.691$ \\

  + Conv.
  & $.405$
  & $.642$
  & $.810$
  & $.733$ \\

  \textbf{+ SimAuthor}
  & $.404$
  & $\mathbf{.747}$
  & $\mathbf{.861}$
  & $\mathbf{.804}$ \\

\bottomrule
\end{tabular*}
\end{minipage}

\end{table}

The authoring objectives are deliberately narrow, raising the possibility that \framework{} merely exploits the feedback function it is given. We therefore test generalization in two complementary ways: by changing the evaluator after authoring, and by changing the objective from distribution matching to downstream classification.
For VSD, AS, COPD, and AF, programs are authored using only the original signal-space objectives and then re-evaluated with frozen pretrained representations that are never exposed to the authoring loop: OPERA-CE for audio \citep{zhang2024opera} and PaPaGei-S for PPG \citep{pillai2025papagei}. Each entry in Table~\ref{tab:generalization}a reports $R_{\mathrm{sel}} \,|\, R_{\max}$, the representation score of the signal-selected program and the best representation score anywhere in its 100-candidate pool. SimAuthor achieves the strongest $R_{\mathrm{sel}}$ on AS, COPD, and AF, and the strongest $R_{\max}$ on VSD and AS. It also exceeds Independent Sampling at the selected endpoint on all four tasks and Text-Opt on three of four. The often substantial gap between $R_{\mathrm{sel}}$ and $R_{\max}$ shows that the authoring and representation evaluators are not interchangeable, while improvement under an unseen representation provides evidence that the search is not merely exploiting the handcrafted authoring
score. 

For LQT and WPW, we test whether authored simulators improve learning from scarce real labels. We compare real-only training with two matched augmentations of 300 positive examples: standard waveform perturbations \citep{raghu2022data} and samples from the canonical SimAuthor program. A frozen ECGFounder representation feeds a class-weighted logistic-regression probe, evaluated on patient-disjoint PTB-XL as in-distribution (ID) and external Chapman--Shaoxing as out-of-distribution (OOD)~\citep{wagner2020ptbxl,zheng2022ecgarrhythmia}.
SimAuthor raises OOD AUPRC from $.633$ to $.747$ for LQT and from $.691$ to $.804$ for WPW, compared with $.642$ and $.733$ under conventional augmentation, while preserving ID performance; OOD AUROC reaches $.889$ and $.965$, respectively.
These results show that the authored ECGs transfer from representation matching to classification of unseen real recordings, with the largest gains under distribution shift. Because authoring and downstream probing both use ECGFounder, we interpret this as task-aligned downstream utility rather than encoder-independent generalization or clinical validity; full details are in \appref{app:ecg-downstream}.

\begin{table}[t]
\centering
\caption{
\textbf{Focused component ablations.}
\textbf{(a)} Final candidate quality, measured by $S_{\max}$ after 100
authoring attempts.
\textbf{(b)} Search efficiency, measured by
$\mathrm{AUC}_{100}$, the mean best-so-far score over the 100-attempt
trajectory.
All values are medians over three seeds.
Parentheses report the change relative to SimAuthor for the same metric.
}
\label{tab:ablation}

\footnotesize
\renewcommand{\arraystretch}{1.06}

\begin{minipage}[t]{0.49\textwidth}
\vspace{0pt}
\centering
\textbf{(a) Final quality: $S_{\max}@100$}\par\vspace{2pt}

\setlength{\tabcolsep}{2.2pt}
\begin{tabular*}{\linewidth}{
@{\extracolsep{\fill}}
lccc
@{}
}
\toprule
\textbf{Task}
& \textbf{SimAuthor}
& \textbf{$-$ Report}
& \textbf{$-$ Library} \\
\midrule

VSD
& .587
& .366 ($-.221$)
& .399 ($-.188$) \\

AS
& .557
& .152 ($-.405$)
& .383 ($-.174$) \\

COPD
& .672
& .609 ($-.063$)
& .662 ($-.010$) \\

AF
& .738
& .589 ($-.148$)
& .719 ($-.019$) \\

LQT
& .568
& .460 ($-.108$)
& .540 ($-.028$) \\

WPW
& .581
& .492 ($-.089$)
& .573 ($-.008$) \\

\bottomrule
\end{tabular*}
\end{minipage}
\hfill
%
\begin{minipage}[t]{0.49\textwidth}
\vspace{0pt}
\centering
\textbf{(b) Search efficiency: $\mathrm{AUC}_{100}$}\par\vspace{2pt}

\setlength{\tabcolsep}{2.2pt}
\begin{tabular*}{\linewidth}{
@{\extracolsep{\fill}}
lccc
@{}
}
\toprule
\textbf{Task}
& \textbf{SimAuthor}
& \textbf{$-$ Report}
& \textbf{$-$ Library} \\
\midrule

VSD
& .395
& .356 ($-.039$)
& .348 ($-.047$) \\

AS
& .458
& .143 ($-.315$)
& .324 ($-.134$) \\

COPD
& .544
& .585 ($+.041$)
& .529 ($-.016$) \\

AF
& .717
& .556 ($-.161$)
& .686 ($-.031$) \\

LQT
& .539
& .439 ($-.101$)
& .514 ($-.026$) \\

WPW
& .574
& .486 ($-.088$)
& .547 ($-.027$) \\

\bottomrule
\end{tabular*}
\end{minipage}

\end{table}

\subsection{Structured feedback and memory support cumulative authoring}
\label{sec:ablation}

We focus on the two components that determine how information from previous attempts is carried into subsequent revisions: the structured discrepancy report and the mechanism library. Table~\ref{tab:ablation} evaluates both final
candidate quality, $S_{\max}@100$, and search efficiency,
$\mathrm{AUC}_{100}$. All values are medians over three seeds;
complete seed-level results are provided in
\appref{app:multiseed-ablation}. Removing the discrepancy report reduces final quality on every task, by $.063$--$.405$, with particularly large losses on AS ($-.405$), VSD ($-.221$), and AF ($-.148$). Search efficiency also decreases on five of six
tasks. This indicates that a scalar score is insufficient for reliable revision: the report is what turns empirical mismatch into directional information about how the retained program should change. COPD is the only
exception in $\mathrm{AUC}_{100}$, where removing the report improves the trajectory average despite lowering the final score, showing that more informative feedback does not necessarily improve every intermediate search step. Removing the mechanism library has a smaller effect on final endpoints ($-.008$ to $-.188$), but lowers $\mathrm{AUC}_{100}$ on all six tasks. The contrast suggests a different role for memory: rather than primarily unlocking isolated high-scoring candidates, the library helps useful implementation mechanisms persist and be reused, making progress accumulate more efficiently across revisions.

\section{Analysis}
\label{sec:analysis}

Three alternative explanations could account for the score gains in
Section~\ref{sec:results}: parameter retuning rather than genuine program authoring, overfitting to the small search-reference set rather than the condition, or sensitivity to the exploration setting or foundation model. We test each in turn.

\subsection{SimAuthor improves simulators through structural revision}
\label{sec:code-analysis}

\begin{figure}[ht]
\centering

\begin{minipage}[t]{0.31\linewidth}
\vspace{0pt}
\begin{saCodePanel}{(a) COPD \hfill $\delta$ +.150}
\saDiffHead{saDiffRed}{--- parent.py}
\begin{saBeforeBlock}
# one cycle for the recording
cycle_len = 60.0 / rr
phase = (t \% cycle_len) / cycle_len
\end{saBeforeBlock}
\vspace{0.8pt}
\saDiffHead{saDiffGreen}{+++ child.py}
\begin{saAfterBlock}
# timing resampled per breath
while idx < len(t):
  t_i = np.random.normal(1.2, .1)
  t_e = np.random.normal(3.0, .3)
  idx += n_breath + n_p
\end{saAfterBlock}
\end{saCodePanel}
\end{minipage}%
\hfill%
\begin{minipage}[t]{0.335\linewidth}
\vspace{0pt}
\begin{saCodePanel}{(b) VSD \hfill $\delta$ +.134}
\saDiffHead{saDiffRed}{--- parent.py}
\begin{saBeforeBlock}
# one spectrum for every case
mur_low = np.random.uniform(80, 150)
mur_high = np.random.uniform(400, 600)
\end{saBeforeBlock}
\vspace{0.8pt}
\saDiffHead{saDiffGreen}{+++ child.py}
\begin{saAfterBlock}
# spectrum set by severity
if severity < .5:
  mur_low = np.random.uniform(100, 200)
  mur_high = np.random.uniform(400, 800)
else: ...  # lower, narrower band
\end{saAfterBlock}
\end{saCodePanel}
\end{minipage}%
\hfill%
\begin{minipage}[t]{0.33\linewidth}
\vspace{0pt}
\begin{saTablePanel}{(c) Revision audit \hfill 18 runs}
\centering

\begin{tikzpicture}[x=1.1cm,y=1.05cm]

\def\xlab{0.48}      
\def\xn{1.12}        
\def\xL{1.28}        
\def\barw{1.52}      
\def\xg{3.58}        

\newcommand{\auditrow}[6]{%
  \pgfmathsetmacro{\sw}{\barw*(#3)/(#4)}%

  \node[
    anchor=east,
    font=#6\fontsize{5.7pt}{6.2pt}\selectfont
  ] at (\xlab,#1) {#2};

  \node[
    anchor=east,
    font=#6\fontsize{4.75pt}{5.2pt}\selectfont,
    text=black!48
  ] at (\xn,#1) {#3/#4};

  \fill[
    black!10,
    rounded corners=0.4pt
  ]
  (\xL,#1-0.067) rectangle +(\barw,0.134);

  \fill[
    black!65,
    rounded corners=0.4pt
  ]
  (\xL,#1-0.067) rectangle +(\sw,0.134);

  \node[
    anchor=east,
    font=#6\fontsize{5.55pt}{6.0pt}\selectfont,
    text=saDiffGreen
  ] at (\xg,#1) {#5\%};
}

\node[
  anchor=east,
  font=\fontsize{4.55pt}{5.0pt}\selectfont,
  text=black!42
] at (\xn,1.94) {$n/N$};

\node[
  anchor=center,
  font=\fontsize{4.55pt}{5.0pt}\selectfont,
  text=black!42
] at (\xL+\barw/2,1.94) {structural share};

\node[
  anchor=east,
  font=\fontsize{4.55pt}{5.0pt}\selectfont,
  text=saDiffGreen
] at (\xg,1.94) {gain};

\auditrow{1.70}{VSD}  {17}{17} {100.0}{}
\auditrow{1.46}{AS}   {13}{16} {85.8} {}
\auditrow{1.22}{COPD} {26}{28} {98.7} {}
\auditrow{0.98}{AF}   {11}{21} {61.2} {}
\auditrow{0.74}{LQT}  {32}{39} {98.8} {}
\auditrow{0.50}{WPW}  {12}{17} {83.9} {}

\fill[
  black!4,
  rounded corners=0.8pt
] (0.05,0.16) rectangle (3.64,0.39);

\auditrow{0.275}{All}{111}{138}{86.1}{\bfseries}

\end{tikzpicture}

\end{saTablePanel}
\end{minipage}

\vspace{-2pt}
\caption{
\textbf{Structural revision during authoring.}
\textbf{(a,b)} Abridged code changes.
\textbf{(c)} Structural share of 138 selected-lineage revisions and their
signed score gain.
}
\label{fig:code-analysis}
\end{figure}
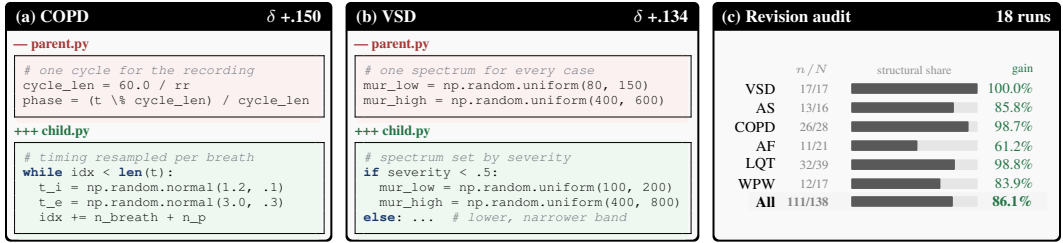

To test whether iterative authoring changes simulator logic rather than merely retuning constants, we trace the selected final simulator in each of the 18 \framework{} runs back to the root and audit every scored parent--child transition along that ancestry path. Of these, 111 are structural and account for 86.1\% of the net score improvement (Figure~\ref{fig:code-analysis}; full audit protocol in \appref{app:audit}).
Structural revisions form the majority on five of six tasks and explain 83.9--100\% of the gain on those tasks. The examples show that these changes introduce new generative logic (e.g. breath-level stochastic timing in COPD and severity-conditioned spectra in VSD) rather than merely adjusting existing
constants. This suggests that cumulative authoring improves simulators primarily by evolving their executable structure.

\subsection{Score gains generalize to held-out recordings}
\label{sec:heldout}

\begin{table}[ht]
  \centering
  \caption{
  \textbf{Generalization to held-out recordings.}
  VSD and AS use patient-disjoint ID splits; the remaining tasks use external
  OOD datasets. Scores use the same evaluator as authoring.
  \framework{} reports the median across three seeds; parentheses show gains
  over the root. See \appref{app:datasets} for dataset details.
  }
  \label{tab:heldout}

  \small
  \ra{1.05}
  \setlength{\tabcolsep}{3.4pt}

  \begin{tabular}{@{}lcccccc@{}}
  \toprule
  &
  \multicolumn{2}{c}{\textit{Patient-disjoint ID}}
  &
  \multicolumn{4}{c}{\textit{External OOD}}
  \\
  \cmidrule(lr){2-3}
  \cmidrule(l){4-7}

  Method & VSD & AS & COPD & AF & LQT & WPW \\
  \midrule

  Root $\mathcal S^{(0)}$
  & .223
  & .103
  & .088
  & .104
  & .277
  & .385 \\

  Sampling
  & .425 {(+.202)}
  & .146 {(+.043)}
  & .362 {(+.275)}
  & .117 {(+.013)}
  & .341 {(+.064)}
  & .443 {(+.059)} \\

  \framework{}
  & \textbf{.577} {(+.354)}
  & \textbf{.440} {(+.338)}
  & \textbf{.507} {(+.420)}
  & \textbf{.204} {(+.100)}
  & \textbf{.480} {(+.203)}
  & \textbf{.538} {(+.153)} \\

  \bottomrule
  \end{tabular}
\end{table}

A high search score could reflect fitting the few reference recordings rather than the underlying condition, so we re-evaluate selected candidates on OOD recordings whenever available. VSD and AS use patient-disjoint in-domain splits, while COPD, AF, LQT, and WPW are evaluated on external OOD datasets (Table~\ref{tab:heldout}; \appref{app:datasets}).  Every final simulator improves on its root, by $+.100$ to $+.420$, and median \framework{} exceeds the authoring-objective-selected best-of-100 sampling candidate on all six conditions, by $+.087$ to $+.294$. Thus the gains transfer to unseen recordings under the authoring evaluator rather than merely fitting the search patients.

\subsection{Robustness to the exploration setting and the foundation model}
\label{sec:robustness}

\begin{table}[ht]
  \centering
  \caption{
  \textbf{Sensitivity to exploration and authoring backbone.}
  Final $S_{\max}@100$ over three seeds. Each cell lists scores in ascending
  order (median in the middle); bold marks the highest median for each task.
  }
  \label{tab:robust}

  \footnotesize
  \renewcommand{\arraystretch}{1.18}

  \begin{minipage}[t]{0.565\linewidth}
  \centering
  \setlength{\tabcolsep}{2.8pt}

  \begin{tabular}{@{}lccc@{}}
  \toprule
  & \multicolumn{3}{c}{\textit{PUCT exploration}} \\
  \cmidrule(lr){2-4}
  Task & $c=.10$ & $c=.25$ & $c=.50$ \\
  \midrule

  VSD
  & {\fontsize{7.7pt}{8.4pt}\selectfont
     [.330, .362, .472]}
  & {\fontsize{7.7pt}{8.4pt}\selectfont
     [.448, \textbf{.587}, .611]}
  & {\fontsize{7.7pt}{8.4pt}\selectfont
     [.488, .504, .535]}
  \\

  LQT
  & {\fontsize{7.7pt}{8.4pt}\selectfont
     [.513, .548, .554]}
  & {\fontsize{7.7pt}{8.4pt}\selectfont
     [.541, .568, .594]}
  & {\fontsize{7.7pt}{8.4pt}\selectfont
     [.572, \textbf{.596}, .637]}
  \\

  \bottomrule
  \end{tabular}
  \end{minipage}%
  \hfill
  \begin{minipage}[t]{0.405\linewidth}
  \centering
  \setlength{\tabcolsep}{3.0pt}

  \begin{tabular}{@{}lcc@{}}
  \toprule
  & \multicolumn{2}{c}{\textit{Authoring backbone}} \\
  \cmidrule(lr){2-3}
  Task & DeepSeek-V4 Pro & Gemini 3.1 Pro \\
  \midrule

  VSD
  & {\fontsize{7.7pt}{8.4pt}\selectfont
     [.543, .551, .579]}
  & {\fontsize{7.7pt}{8.4pt}\selectfont
     [.448, \textbf{.587}, .611]}
  \\

  LQT
  & {\fontsize{7.7pt}{8.4pt}\selectfont
     [.575, \textbf{.598}, .614]}
  & {\fontsize{7.7pt}{8.4pt}\selectfont
     [.541, .568, .594]}
  \\

  \bottomrule
  \end{tabular}
  \end{minipage}
\end{table}

We repeat VSD and LQT authoring with three exploration coefficients and two strong models from different families, Gemini 3.1 Pro~\citep{deepmind2026gemini31pro} and open-weight DeepSeek-V4 Pro~\citep{deepseek2026v4}, using three seeds per setting.
The three-seed results show task-dependent sensitivity rather than a universally optimal configuration. For exploration, $c=.25$ gives the highest median on VSD (.587), whereas $c=.50$ performs best on LQT (.596). Backbone choice shows a similar crossover: Gemini 3.1 Pro has the higher VSD median (.587 vs.\ .551), while DeepSeek-V4 Pro has the higher LQT median (.598 vs.\ .568). Thus, the observed gains are not specific to a single search setting or foundation-model backbone, although their relative effectiveness depends on the task.

\section{Related Work}
\paragraph{LM-guided program search and scientific software.}
Foundation models are increasingly embedded in execution-and-feedback loops that iteratively improve programs, with applications spanning mathematics, algorithm discovery, reward design, symbolic regression, and scientific software
\citep{romera2024mathematical,novikov2025alphaevolve,ma2023eureka,
shojaee2024llmsr,ma2024llm,aygun2025ai,lehman2023elm}.
Related work on test-time scaling likewise studies
how inference budgets should be divided between independent sampling and iterative refinement \citep{brown2024monkeys,snell2024scaling}; our focus is how such refinement can accumulate when the feedback itself is only a weak proxy for artifact quality.

\paragraph{Agentic refinement and persistent state.}
Iterative agents improve outputs through repeated critique and revision~\citep{madaan2023selfrefine,shinn2023reflexion}, while other systems retain state across iterations in the form of skills, strategies, reusable code, or
heuristics~\citep{wang2023voyager,lehman2023elm,ye2024reevo}.
Prompt and textual optimization methods provide another form of persistent state, in which the object being updated is a natural-language prompt or strategy~\citep{yang2024opro,agrawal2025gepa}.
\framework{} combines iterative refinement with artifact-centered persistence: the executable simulator itself is inherited and revised across calls, while a mechanism library provides additional reusable implementation knowledge.

\paragraph{Scientific simulation.}
Classical biomedical simulators encode explicit generative mechanisms with interpretable parameters, as in established ECG and PPG models~\citep{mcsharry2003dynamical,solosenko2017ppg}. Learned generators such as GANs and diffusion models instead approximate biomedical or audio signal distributions implicitly~\citep{golany2020simgans,kong2020diffwave,liu2023audioldm}.
Existing approaches therefore largely either hand-design an explicit simulator or learn an implicit generator from data. We study a complementary setting:
a foundation model authors and iteratively refines the explicit, executable simulator itself from weak empirical feedback. The resulting artifact is an inspectable generative model rather than only a source of synthetic samples.

\section{Conclusion and Discussion}

This work establishes scientific simulator authoring as a viable use of foundation models under weak empirical feedback. Across six biomedical tasks, \framework{} generally outperforms independent sampling, PUCT score search, and textual-strategy optimization; its gains also persist on unseen recordings,
alternative representations, and downstream ECG evaluation. Together with the revision audit, these results suggest that foundation-model knowledge can be progressively converted into better executable scientific simulators.

\paragraph{Limitations and future directions.}
Our evaluation focuses primarily on \emph{intrinsic} agreement between simulated and observed signals, which does not establish physiological or mechanistic fidelity. Establishing such fidelity would also enable uses beyond distribution matching. In the forward direction, explicit simulator variables
could be set to represent rare conditions or prescribed parameter regimes, producing targeted synthetic data under controlled assumptions \citep{vanbreugel2024synthetic}. In the reverse direction, simulator parameters or latent mechanisms could be inferred by searching for settings whose simulated
observations best match a given real recording, connecting authored simulators to simulation-based inference \citep{cranmer2020frontier}. Once sufficiently validated and calibrated to an individual, the same simulator could be perturbed to study how changes in specific mechanisms or interventions affect predicted observations, providing
a possible route toward digital-twin applications
\citep{bjornsson2020digitaltwins,corralacero2020digitaltwin}. These capabilities are outside the present study; evaluating such \emph{extrinsic} utility is an important direction for future work.

Authoring performance also depends on the evaluator: the model can only improve discrepancies that the feedback exposes. This may be particularly important in modalities where clinically relevant morphology is difficult to summarize with
generic signal statistics, and may partly explain the weaker advantage of persistent authoring on AF. Future work should incorporate expert-designed features, physiological constraints, and complementary evaluators, and systematically study how feedback quality affects authoring. Our empirical
scope is also limited to three seeds per setting and predominantly strong frontier models; testing smaller or less capable models would help characterize how persistent authoring interacts with foundation-model capability.

Finally, the framework is not specific to biomedical signals. The same setting arises wherever an executable simulator can be compared with observations, including industrial sensing and physical systems. Extending \framework{} to such domains, with domain-specific evaluation and stronger expert involvement,
will test the broader generality of scientific simulator authoring.

\section*{Acknowledgments}
\small{This work was supported by the NWO AiNed Fellowship Grant of A.S., and in part by Google.org and the Google Cloud Research Credits program through the Gemini Academic Program. We also acknowledge the use of the Dutch National Supercomputer Snellius for essential computational tasks.}

\section*{Reproducibility Statement}

The appendices report dataset identities and splits, evaluator definitions, the end-to-end authoring algorithm, complete prompts, numerical settings, multi-seed results, compute and token-matched controls, downstream ECG evaluation, and the revision-audit procedure. The code for executing and evaluating simulators, all authored simulator programs of the main runs, and the search traces needed to reconstruct the reported trajectories are available at \url{https://github.com/yishani/SimAuthor}.

\section*{Ethics Statement}

All biomedical datasets used in this study are publicly released,
de-identified research datasets used under their original licenses; no new human-subject data were collected. The authored simulators and synthetic signals are research artifacts and may inherit biases or incidental properties of the source data and evaluators. We make no diagnostic or clinical claims.
Physiological validation, patient-specific use, or applications such as intervention modeling would require independent expert and prospective validation beyond the scope of this work.

\section*{AI Use Statement}

Foundation models are experimental components of the method rather than auxiliary tools. Canonical \framework{} runs use Gemini 3.1 Pro for the authoring pipeline, including generation of the scientific blueprint and executable simulators; the backbone robustness experiment additionally uses
DeepSeek-V4 Pro. All authoring prompts are reproduced in
\appref{app:prompts}. A separate DeepSeek-V4-Flash model is used only in the revision audit to help classify program changes after deterministic diff processing and to assess mechanism reuse. These annotations affect the audit analysis, but not the authoring trajectories, evaluator scores, or downstream results.

\newpage

\bibliographystyle{simauthor_arxiv}
\bibliography{references}

@article{altan2017multimedia,
  title={Multimedia respiratory database (RespiratoryDatabase@ TR): Auscultation sounds and chest X-rays},
  author={Altan, G{\"o}khan and Kutlu, Yakup and Garbi, Yusuf and Pekmezci, Adnan {\"O}zhan and Nural, Serkan},
  journal={Natural and Engineering Sciences},
  volume={2},
  number={3},
  pages={59--72},
  year={2017},
  publisher={Cemal TURAN}
}

@article{aygun2025ai,
  title={An AI system to help scientists write expert-level empirical software},
  author={Ayg{\"u}n, Eser and Belyaeva, Anastasiya and Comanici, Gheorghe and Coram, Marc and Cui, Hao and Garrison, Jake and Johnston, Renee and Kast, Anton and McLean, Cory Y and Norgaard, Peter and others},
  journal={arXiv preprint arXiv:2509.06503},
  year={2025}
}

@article{shojaee2024llmsr,
  title={Llm-sr: Scientific equation discovery via programming with large language models},
  author={Shojaee, Parshin and Meidani, Kazem and Gupta, Shashank and Farimani, Amir Barati and Reddy, Chandan K},
  journal={arXiv preprint arXiv:2404.18400},
  year={2024}
}

@article{charlton2022detecting,
  title={Detecting beats in the photoplethysmogram: benchmarking open-source algorithms},
  author={Charlton, Peter H and Kotzen, Kevin and Mej{\'\i}a-Mej{\'\i}a, Elisa and Aston, Philip J and Budidha, Karthik and Mant, Jonathan and Pettit, Callum and Behar, Joachim A and Kyriacou, Panicos A},
  journal={Physiological Measurement},
  volume={43},
  number={8},
  pages={085007},
  year={2022},
  publisher={IOP Publishing}
}

@article{chen2021evaluating,
  title={Evaluating large language models trained on code},
  author={Chen, Mark and Tworek, Jerry and Jun, Heewoo and Yuan, Qiming and Pinto, Henrique Ponde De Oliveira and Kaplan, Jared and Edwards, Harri and Burda, Yuri and Joseph, Nicholas and Brockman, Greg and others},
  journal={arXiv preprint arXiv:2107.03374},
  year={2021}
}

@article{fraiwan2021dataset,
  title={A dataset of lung sounds recorded from the chest wall using an electronic stethoscope},
  author={Fraiwan, Mohammad and Fraiwan, Luay and Khassawneh, Basheer and Ibnian, Ali},
  journal={Data in Brief},
  volume={35},
  pages={106913},
  year={2021},
  publisher={Elsevier}
}

@inproceedings{golany2020simgans,
  title={SimGANs: Simulator-based generative adversarial networks for ECG synthesis to improve deep ECG classification},
  author={Golany, Tomer and Radinsky, Kira and Freedman, Daniel},
  booktitle={International Conference on Machine Learning},
  pages={3597--3606},
  year={2020},
  organization={PMLR}
}

@article{guo2024deepseek,
  title={DeepSeek-Coder: when the large language model meets programming--the rise of code intelligence},
  author={Guo, Daya and Zhu, Qihao and Yang, Dejian and Xie, Zhenda and Dong, Kai and Zhang, Wentao and Chen, Guanting and Bi, Xiao and Wu, Yifan and Li, YK and others},
  journal={arXiv preprint arXiv:2401.14196},
  year={2024}
}

@article{jia2024zchsound,
  author={Jia, Weijie and Wang, Yunyan and Chen, Renwei and Ye, Jingjing and Li, Die and Yin, Fei and Yu, Jin and Chen, Jiajia and Shu, Qiang and Xu, Weize},
  journal={IEEE Transactions on Biomedical Engineering},
  title={{ZCHSound}: Open-Source {ZJU} Paediatric Heart Sound Database With Congenital Heart Disease},
  year={2024},
  volume={71},
  number={8},
  pages={2278--2286},
  doi={10.1109/TBME.2023.3348800}
}

@article{kong2020diffwave,
  title={Diffwave: A versatile diffusion model for audio synthesis},
  author={Kong, Zhifeng and Ping, Wei and Huang, Jiaji and Zhao, Kexin and Catanzaro, Bryan},
  journal={arXiv preprint arXiv:2009.09761},
  year={2020}
}

@incollection{lehman2023elm,
  title     = {Evolution Through Large Models},
  author    = {Lehman, Joel and Gordon, Jonathan and Jain, Shawn and Ndousse, Kamal and Yeh, Cathy and Stanley, Kenneth O.},
  booktitle = {Handbook of Evolutionary Machine Learning},
  editor    = {Banzhaf, Wolfgang and Machado, Penousal and Zhang, Mengjie},
  publisher = {Springer Nature},
  address   = {Singapore},
  pages     = {331--366},
  year      = {2023},
  doi       = {10.1007/978-981-99-3814-8_11}
}

@article{li2023starcoder,
  title={Starcoder: may the source be with you!},
  author={Li, Raymond and Allal, Loubna Ben and Zi, Yangtian and Muennighoff, Niklas and Kocetkov, Denis and Mou, Chenghao and Marone, Marc and Akiki, Christopher and Li, Jia and Chim, Jenny and others},
  journal={arXiv preprint arXiv:2305.06161},
  year={2023}
}

@article{liu2023audioldm,
  title={Audioldm: Text-to-audio generation with latent diffusion models},
  author={Liu, Haohe and Chen, Zehua and Yuan, Yi and Mei, Xinhao and Liu, Xubo and Mandic, Danilo and Wang, Wenwu and Plumbley, Mark D},
  journal={arXiv preprint arXiv:2301.12503},
  year={2023}
}

@article{ma2023eureka,
  title={Eureka: Human-level reward design via coding large language models},
  author={Ma, Yecheng Jason and Liang, William and Wang, Guanzhi and Huang, De-An and Bastani, Osbert and Jayaraman, Dinesh and Zhu, Yuke and Fan, Linxi and Anandkumar, Anima},
  journal={arXiv preprint arXiv:2310.12931},
  year={2023}
}

@article{ma2024llm,
  title={Llm and simulation as bilevel optimizers: A new paradigm to advance physical scientific discovery},
  author={Ma, Pingchuan and Wang, Tsun-Hsuan and Guo, Minghao and Sun, Zhiqing and Tenenbaum, Joshua B and Rus, Daniela and Gan, Chuang and Matusik, Wojciech},
  journal={arXiv preprint arXiv:2405.09783},
  year={2024}
}

@article{mcsharry2003dynamical,
  title={A dynamical model for generating synthetic electrocardiogram signals},
  author={McSharry, Patrick E and Clifford, Gari D and Tarassenko, Lionel and Smith, Leonard A},
  journal={IEEE transactions on biomedical engineering},
  volume={50},
  number={3},
  pages={289--294},
  year={2003},
  publisher={IEEE}
}

@article{novikov2025alphaevolve,
  title={Alphaevolve: A coding agent for scientific and algorithmic discovery},
  author={Novikov, Alexander and V{\~u}, Ng{\^a}n and Eisenberger, Marvin and Dupont, Emilien and Huang, Po-Sen and Wagner, Adam Zsolt and Shirobokov, Sergey and Kozlovskii, Borislav and Ruiz, Francisco JR and Mehrabian, Abbas and others},
  journal={arXiv preprint arXiv:2506.13131},
  year={2025}
}

@inproceedings{rocha2017respiratory,
  title={A respiratory sound database for the development of automated classification},
  author={Rocha, BM and Filos, Dimitris and Mendes, Lea and Vogiatzis, Ioannis and Perantoni, Eleni and Kaimakamis, Evangelos and Natsiavas, P and Oliveira, Ana and J{\'a}come, C and Marques, A and others},
  booktitle={International Conference on Biomedical and Health Informatics},
  pages={33--37},
  year={2017},
  organization={Springer}
}

@article{romera2024mathematical,
  title={Mathematical discoveries from program search with large language models},
  author={Romera-Paredes, Bernardino and Barekatain, Mohammadamin and Novikov, Alexander and Balog, Matej and Kumar, M Pawan and Dupont, Emilien and Ruiz, Francisco JR and Ellenberg, Jordan S and Wang, Pengming and Fawzi, Omar and others},
  journal={Nature},
  volume={625},
  number={7995},
  pages={468--475},
  year={2024},
  publisher={Nature Publishing Group UK London}
}

@article{shinn2023reflexion,
  title={Reflexion: Language agents with verbal reinforcement learning},
  author={Shinn, Noah and Cassano, Federico and Gopinath, Ashwin and Narasimhan, Karthik and Yao, Shunyu},
  journal={Advances in neural information processing systems},
  volume={36},
  pages={8634--8652},
  year={2023}
}

@article{wagner2020ptbxl,
  title={PTB-XL, a large publicly available electrocardiography dataset},
  author={Wagner, Patrick and Strodthoff, Nils and Bousseljot, Ralf-Dieter and Kreiseler, Dieter and Lunze, Fatima I and Samek, Wojciech and Schaeffter, Tobias},
  journal={Scientific data},
  volume={7},
  number={1},
  pages={154},
  year={2020}
}

@article{wang2023voyager,
  title={Voyager: An open-ended embodied agent with large language models},
  author={Wang, Guanzhi and Xie, Yuqi and Jiang, Yunfan and Mandlekar, Ajay and Xiao, Chaowei and Zhu, Yuke and Fan, Linxi and Anandkumar, Anima},
  journal={arXiv preprint arXiv:2305.16291},
  year={2023}
}

@inproceedings{ye2024reevo,
  title={{ReEvo}: Large Language Models as Hyper-Heuristics with Reflective Evolution},
  author={Ye, Haoran and Wang, Jiarui and Cao, Zhiguang and Berto, Federico and Hua, Chuanbo and Kim, Haeyeon and Park, Jinkyoo and Song, Guojie},
  booktitle={NeurIPS},
  year={2024}
}

@article{rieke2020federated,
  title     = {The future of digital health with federated learning},
  author    = {Rieke, Nicola and Hancox, Jonny and Li, Wenqi and Milletar{\`i}, Fausto and Roth, Holger R and Albarqouni, Shadi and Bakas, Spyridon and Galtier, Mathieu N and Landman, Bennett A and Maier-Hein, Klaus and Ourselin, S{\'e}bastien and Sheller, Micah J and Summers, Ronald M and Trask, Andrew and Xu, Daguang and Baust, Maximilian and Cardoso, M Jorge},
  journal   = {npj Digital Medicine},
  volume    = {3},
  number    = {1},
  pages     = {119},
  year      = {2020},
  publisher = {Nature Publishing Group},
  doi       = {10.1038/s41746-020-00323-1}
}

@article{ali2026buet,
  title={Buet multi-disease heart sound dataset: A comprehensive auscultation dataset for developing computer-aided diagnostic systems},
  author={Ali, Shams Nafisa and Zahin, Afia and Shuvo, Samiul Based and Nizam, Nusrat Binta and Nuhash, Shoyad Ibn Sabur Khan and Razin, Sayeed Sajjad and Sani, SM Sakeef and Rahman, Farihin and Nizam, Nawshad Binta and Azam, Farhat Binte and others},
  journal={Computer Methods and Programs in Biomedicine Update},
  pages={100237},
  year={2026},
  publisher={Elsevier}
}

@article{singhal2023large,
  title     = {Large language models encode clinical knowledge},
  author    = {Singhal, Karan and Azizi, Shekoofeh and Tu, Tao and Mahdavi, S. Sara and Wei, Jason and Chung, Hyung Won and Scales, Nathan and Tanwani, Ajay and Cole-Lewis, Heather and Pfohl, Stephen and Payne, Perry and Seneviratne, Martin and Gamble, Paul and Kelly, Chris and Scharli, Nathanael and Chowdhery, Aakanksha and Mansfield, Philip and Aguera y Arcas, Blaise and Webster, Dale and Corrado, Greg S. and Matias, Yossi and Chou, Katherine and Gottweis, Juraj and Tomasev, Nenad and Karthikesalingam, Alan and Natarajan, Vivek},
  journal   = {Nature},
  volume    = {620},
  pages     = {172--180},
  year      = {2023},
  doi       = {10.1038/s41586-023-06291-2}
}

@misc{bacevicius2024longterm,
  title={Long-term electrocardiogram and wrist-based photoplethysmogram recordings with annotated atrial fibrillation episodes},
  author={Bacevi{\v{c}}ius, Justinas and Plu{\v{s}}{\v{c}}iauskait{\.e}, Vilma and Abramikas, {\v{Z}}ygimantas and Badaras, Ignas and Butkuvien{\.e}, Monika and Daukantas, Saulius and Dvinelis, Ernestas and Gudauskas, Modestas and Jukna, Edvardas and Kiseli{\=u}t{\.e}, Margarita and Kundelis, Ri{\v{c}}ardas and Marinskien{\.e}, Julija and Paliakait{\.e}, Birut{\.e} and Petr{\.e}nas, Andrius and Petrylait{\.e}, Marija and Pilkien{\.e}, Aist{\.e} and Pudinskait{\.e}, Gintar{\.e} and Radavi{\v{c}}ius, Vytautas and Rapalis, Andrius and Sokas, Daivaras and Solo{\v{s}}enko, Andrius and Staigyt{\.e}, Justina and Stankevi{\v{c}}i{\=u}t{\.e}, Guost{\.e} and Taparauskait{\.e}, Neringa and Zarembait{\.e}, Gintar{\.e} and Aidietis, Audrius and Marozas, Vaidotas},
  year={2024},
  publisher={Zenodo},
  doi={10.5281/zenodo.11242869}
}

@article{brown2024monkeys,
  title={Large Language Monkeys: Scaling Inference Compute with Repeated Sampling},
  author={Brown, Bradley and Juravsky, Jordan and Ehrlich, Ryan and Clark, Ronald and Le, Quoc V and R{\'e}, Christopher and Mirhoseini, Azalia},
  journal={arXiv preprint arXiv:2407.21787},
  year={2024}
}

@article{snell2024scaling,
  title={Scaling {LLM} Test-Time Compute Optimally can be More Effective than Scaling Model Parameters},
  author={Snell, Charlie and Lee, Jaehoon and Xu, Kelvin and Kumar, Aviral},
  journal={arXiv preprint arXiv:2408.03314},
  year={2024}
}

@inproceedings{madaan2023selfrefine,
  title={Self-Refine: Iterative Refinement with Self-Feedback},
  author={Madaan, Aman and Tandon, Niket and Gupta, Prakhar and Hallinan, Skyler and Gao, Luyu and Wiegreffe, Sarah and Alon, Uri and Dziri, Nouha and Prabhumoye, Shrimai and Yang, Yiming and others},
  booktitle={Advances in Neural Information Processing Systems},
  volume={36},
  year={2023}
}

@misc{deepmind2026gemini31pro,
  author={{Google DeepMind}},
  title={{Gemini 3.1 Pro} Model Card},
  year={2026},
  howpublished={\url{https://deepmind.google/models/model-cards/gemini-3-1-pro/}}
}

@misc{deepseek2026v4,
  title         = {DeepSeek-V4: Towards Highly Efficient Million-Token Context Intelligence},
  author        = {{DeepSeek-AI} and Xu, Anyi and Lin, Bangcai and Xue, Bing and others},
  year          = {2026},
  eprint        = {2606.19348},
  archivePrefix = {arXiv},
  primaryClass  = {cs.CL},
  url           = {https://arxiv.org/abs/2606.19348}
}

@inproceedings{yang2024opro,
    title     = {Large Language Models as Optimizers},
    author    = {Yang, Chengrun and Wang, Xuezhi and Lu, Yifeng and
                 Liu, Hanxiao and Le, Quoc V. and Zhou, Denny and Chen, Xinyun},
    booktitle = {International Conference on Learning Representations},
    year      = {2024},
    url       = {https://openreview.net/forum?id=Bb4VGOWELI}
  }

@article{agrawal2025gepa,
    title         = {{GEPA}: Reflective Prompt Evolution Can Outperform
                     Reinforcement Learning},
    author        = {Agrawal, Lakshya A. and Tan, Shangyin and Soylu, Dilara and
                     Ziems, Noah and Khare, Rishi and Opsahl-Ong, Krista and
                     Singhvi, Arnav and Shandilya, Herumb and Ryan, Michael J.
                     and Jiang, Meng and Potts, Christopher and Sen, Koushik and
                     Dimakis, Alexandros G. and Stoica, Ion and Klein, Dan and
                     Zaharia, Matei and Khattab, Omar},
    journal       = {arXiv preprint arXiv:2507.19457},
    year          = {2025},
    eprint        = {2507.19457},
    archivePrefix = {arXiv},
    primaryClass  = {cs.LG},
    url           = {https://arxiv.org/abs/2507.19457}
  }

@inproceedings{raghu2022data,
    title     = {Data Augmentation for Electrocardiograms},
    author    = {Raghu, Aniruddh and Shanmugam, Divya and
                 Pomerantsev, Eugene and Guttag, John and
                 Stultz, Collin M.},
    booktitle = {Proceedings of the Conference on Health,
                 Inference, and Learning},
    series    = {Proceedings of Machine Learning Research},
    volume    = {174},
    pages     = {282--310},
    year      = {2022},
    publisher = {PMLR}
  }

@inproceedings{zhang2024opera,
    title     = {Towards Open Respiratory Acoustic Foundation Models:
                 Pretraining and Benchmarking},
    author    = {Zhang, Yuwei and Xia, Tong and Han, Jing and Wu, Yu and
                 Rizos, Georgios and Liu, Yang and Mosuily, Mohammed and
                 Chauhan, Jagmohan and Mascolo, Cecilia},
    booktitle = {Advances in Neural Information Processing Systems},
    volume    = {37},
    year      = {2024}
  }

@inproceedings{pillai2025papagei,
    title     = {{PaPaGei}: Open Foundation Models for Optical
                 Physiological Signals},
    author    = {Pillai, Arvind and Spathis, Dimitris and Kawsar, Fahim and
                 Malekzadeh, Mohammad},
    booktitle = {International Conference on Learning Representations},
    year      = {2025},
    url       = {https://openreview.net/forum?id=qQebH9B2oT}
  }

@article{li2025ecgfounder,
    title     = {An Electrocardiogram Foundation Model Built on over 10 Million Recordings},
    author    = {Li, Jun and Aguirre, Aaron D. and Moura Junior, Valdery
                 and Jin, Jiarui and Liu, Che and Zhong, Lanhai
                 and Sun, Chenxi and Clifford, Gari D.
                 and Westover, M. Brandon and Hong, Shenda},
    journal   = {NEJM AI},
    volume    = {2},
    number    = {7},
    pages     = {AIoa2401033},
    year      = {2025},
    publisher = {Massachusetts Medical Society},
    doi       = {10.1056/AIoa2401033},
    url       = {https://doi.org/10.1056/AIoa2401033}
  }

@article{vanbreugel2024synthetic,
  title   = {Synthetic data in biomedicine via generative artificial intelligence},
  author  = {van Breugel, Boris and Liu, Tennison and Oglic, Dino and van der Schaar, Mihaela},
  journal = {Nature Reviews Bioengineering},
  volume  = {2},
  pages   = {991--1004},
  year    = {2024},
  doi     = {10.1038/s44222-024-00245-7}
}

@article{cranmer2020frontier,
  title   = {The frontier of simulation-based inference},
  author  = {Cranmer, Kyle and Brehmer, Johann and Louppe, Gilles},
  journal = {Proceedings of the National Academy of Sciences},
  volume  = {117},
  number  = {48},
  pages   = {30055--30062},
  year    = {2020},
  doi     = {10.1073/pnas.1912789117}
}

@article{corralacero2020digitaltwin,
  title   = {The `Digital Twin' to enable the vision of precision cardiology},
  author  = {Corral-Acero, Jorge and Margara, Francesca and Marciniak, Maciej
             and Rodero, Cristobal and Loncaric, Filip and Feng, Yingjing
             and Gilbert, Andrew and Fernandes, Joao F. and Bukhari, Hassaan A.
             and Wajdan, Ali and others},
  journal = {European Heart Journal},
  volume  = {41},
  number  = {48},
  pages   = {4556--4564},
  year    = {2020},
  doi     = {10.1093/eurheartj/ehaa159}
}

@article{bjornsson2020digitaltwins,
  title   = {Digital twins to personalize medicine},
  author  = {Bj{\"o}rnsson, Bergthor and Borrebaeck, Carl and Elander, Nils
             and Gasslander, Thomas and Gawel, Danuta R. and Gustafsson, Mika
             and J{\"o}rnsten, Rebecka and Lee, Eun Jung and Li, Xinxiu
             and Lilja, Sandra and others},
  journal = {Genome Medicine},
  volume  = {12},
  pages   = {4},
  year    = {2020},
  doi     = {10.1186/s13073-019-0701-3}
}

@article{solosenko2017ppg,
  title   = {Modeling of the photoplethysmogram during atrial fibrillation},
  author  = {Solo{\v{s}}enko, Andrius and Petr{\.e}nas, Andrius and Marozas, Vaidotas and S{\"o}rnmo, Leif},
  journal = {Computers in Biology and Medicine},
  volume  = {81},
  pages   = {130--138},
  year    = {2017},
  doi     = {10.1016/j.compbiomed.2016.12.016}
}

@article{zheng2022ecgarrhythmia,
  author  = {Zheng, Jianwei and Guo, Hangyuan and Chu, Huimin},
  title   = {A large scale 12-lead electrocardiogram database for arrhythmia study},
  journal = {PhysioNet},
  year    = {2022},
  month   = aug,
  note    = {Version 1.0.0},
  doi     = {10.13026/wgex-er52},
  url     = {https://doi.org/10.13026/wgex-er52}
}

\newpage
\appendix

\newpage
  \section{Evaluator Definitions}
  \label{app:evaluator}

  \paragraph{Common interface.}
  Each evaluator compares a set of generated recordings with a fixed reference
  set and returns a score in $[0,1]$ together with a structured discrepancy
  report. The evaluator and all of its parameters are fixed before search.
  Candidates that fail to execute or violate the output contract are discarded
  before evaluation.

  \paragraph{Audio: acoustic-feature overlap.}
  Recordings are resampled to 16\,kHz and peak-normalized. From each recording we
  extract the frame-averaged values of 13 MFCCs and the zero-crossing rate (ZCR),
  giving 14 features. Each feature is standardized using the reference-set mean
  and standard deviation. For feature $k$, we construct reference and generated
  histograms over shared bin edges and compute their overlap
  \[
  o_k = \sum_b \min\!\left(
      \hat p^{\mathrm{ref}}_{kb},
      \hat p^{\mathrm{gen}}_{kb}
  \right).
  \]
  The final score is
  \[
  s_{\mathrm{audio}}=\frac{1}{14}\sum_{k=1}^{14}o_k.
  \]

  \paragraph{PPG: pulse-level feature similarity.}
  Signals are processed at 125\,Hz using their first 30\,s. After a
  0.5--10\,Hz Butterworth bandpass, systolic peaks are detected with a minimum
  spacing of 300\,ms. We extract ten segment-level features: heart rate,
  pulse-interval mean and standard deviation, RMSSD, systolic-amplitude mean and
  standard deviation, amplitude range, 25--75\% rise time, pulse width at half
  amplitude, and pulse count. Non-physiological intervals outside
  300--2000\,ms are removed.

  For each feature $k$, similarity is computed from the difference between the
  reference and generated means:
  \[
  q_k =
  \exp\!\left[
  -\,\frac{
      2\left|\mu_k^{\mathrm{ref}}-\mu_k^{\mathrm{gen}}\right|
  }{
      \sqrt{\bigl((\sigma_k^{\mathrm{ref}})^2+
                   (\sigma_k^{\mathrm{gen}})^2\bigr)/2}
      +\epsilon
  }
  \right],
  \qquad
  s_{\mathrm{PPG}}=\frac{1}{10}\sum_{k=1}^{10}q_k.
  \]
  After bandpass filtering, systolic peaks are detected using a fixed amplitude threshold and a minimum spacing of 300\,ms.

  \paragraph{ECG: frozen representation-distribution similarity.}
  Each generated example contains two 10\,s leads at 500\,Hz. Lead~I and lead~II
  are placed in the corresponding channels of a 12-lead input, with the remaining
  channels set to zero. After high-pass filtering and per-lead standardization, a
  frozen ECGFounder encoder maps each recording to a 1,024-dimensional embedding
  \citep{li2025ecgfounder}. 

  Let $\{x_i\}_{i=1}^{n}$ and $\{y_j\}_{j=1}^{m}$ denote the reference and
  generated embeddings. We compute the unbiased energy distance
  \[
  D_{\mathrm{E}} =
  \frac{2}{nm}\sum_{i,j}\lVert x_i-y_j\rVert_2
  -\frac{1}{n(n-1)}\sum_{i\ne i'}\lVert x_i-x_{i'}\rVert_2
  -\frac{1}{m(m-1)}\sum_{j\ne j'}\lVert y_j-y_{j'}\rVert_2 ,
  \]
  and transform it into a bounded search score:
  \[
  s_{\mathrm{ECG}}
  =
  \frac{1}{
  1+\max(D_{\mathrm{E}},0)/0.15
  }.
  \]

  \paragraph{Discrepancy reports.}
  The report makes the fixed objective interpretable to the refiner. For audio it
  lists each feature's overlap, standardized mean shift, and generated-to-reference
  spread ratio. For PPG it reports each feature's similarity and the corresponding
  means in physical units. For ECG it reports energy distance, centroid
  similarity, spread ratio, and within-set similarity as a mode-collapse
  diagnostic. In the full system, this report and a rendered comparison of three
  reference and three generated recordings are provided to the refiner.

\newpage
\section{End-to-End Authoring Loop}\label{app:algorithm}

\begin{algorithm}[H]
\small
\caption{End-to-end simulator authoring with \framework}
\label{alg:simauthor}
\begin{algorithmic}[1]
\Require condition--modality task, search-reference set $\mathcal D_{\mathrm{ref}}$, evaluator, budget $T=100$, extraction threshold $\delta$
\State $B \gets$ \textsc{Blueprint}() \Comment{one model call; persistent context}
\State $\mathcal S^{(0)} \gets$ \textsc{RootSimulator}() \Comment{one independent model call}
\State $\mathcal T \gets \{\mathcal S^{(0)}\}$; \quad $M \gets \varnothing$ \Comment{candidate pool, mechanism library}
\State execute $\mathcal S^{(0)}$, score against $\mathcal D_{\mathrm{ref}}$
\For{$t = 1, \dots, T$}
  \State $n_p \gets \arg\max_{n \in \mathcal T}\big[\tilde v(n) + c_{\mathrm{puct}}\,p(n)\sqrt{N_{\mathrm{tot}}}/(1+N(n))\big]$ \Comment{select parent}
  \State $c \gets \big(B,\ \text{code}(n_p),\ \text{report}(n_p),\ \text{visual}(n_p),\ \text{last } k \text{ entries of } M\big)$ \Comment{assemble context}
  \State $\mathcal S_{\mathrm{child}} \gets$ \textsc{Refine}$(c)$ \Comment{one model call: one focused revision}
  \If{$\mathcal S_{\mathrm{child}}$ executes and satisfies the modality output contract}
    \State execute $\mathcal S_{\mathrm{child}}$; score and produce a discrepancy report against $\mathcal D_{\mathrm{ref}}$
    \State $\mathcal T \gets \mathcal T \cup \{\mathcal S_{\mathrm{child}}\}$ with a link to $n_p$
    \If{$\mathrm{score}(\mathcal S_{\mathrm{child}}) - \mathrm{score}(n_p) \geq \delta$}
      \State $M \gets M \cup \{\textsc{ExtractMechanism}(n_p, \mathcal S_{\mathrm{child}})\}$ \Comment{one model call}
    \EndIf
  \EndIf \Comment{invalid children consume the attempt}
\EndFor
\State \Return best-so-far candidate at each budget
\end{algorithmic}
\end{algorithm}

\newpage
  \section{Mechanism Library}
  \label{app:mechlib}

  The mechanism library is a compact program-level memory of productive
  revisions. When a valid child program $S_{n+1}$ improves its selected parent
  $S_n$ by at least $\delta$, an extraction call summarizes the program diff
  $S_n\!\rightarrow S_{n+1}$ as a normalized mechanism name, a behavioral
  description, and a representative code fragment. The score improvement and
  source revision are retained as provenance. Entries with the same normalized
  name are merged by incrementing their observation count. 
 Before each refinement call, up to eight recent mechanisms are included as
  optional implementation suggestions. The refiner is instructed to reuse or
  adapt an entry only when it is relevant to the current parent program and
  discrepancy report. The library therefore transfers implementable discoveries
  across iterations without forcing every later program to contain every stored
  mechanism.

  For example, one VSD revision replaced a severity-independent murmur
  distribution with a severity-conditioned generator. The child improved its
  parent from $.3631$ to $.4971$ ($\Delta=+.1341$), producing the following
  library entry:

  \begin{tcolorbox}[
    colframe=black,
    title=\small{Example mechanism-library entry}
  ]
  \small

  \textbf{Name.}\quad
  \texttt{severity\_dependent\_murmur\_spectrum}

  \medskip
  \textbf{Behavior.}\quad
  Vary the murmur frequency band and resonance with disease severity to represent
  the acoustic differences between restrictive and unrestrictive ventricular
  septal defects.

  \medskip
  \textbf{Representative code.}
  \begin{lstlisting}
  if severity < 0.5:
      murmur_low = np.random.uniform(100, 200)
      resonance = np.random.uniform(300, 500)
  else:
      murmur_low = np.random.uniform(50, 100)
      resonance = np.random.uniform(150, 250)
  \end{lstlisting}

  \textbf{Provenance.}\quad
  One parent--child revision,
  $.3631\!\rightarrow\!.4971$ ($\Delta=+.1341$).

  \end{tcolorbox}

  A later refiner receives the entry under a \emph{Reusable Mechanisms} heading,
  together with the instruction that it may reuse, modify, or ignore the
  mechanism. The exact extraction prompt, storage format, and complete per-run
  libraries are released with the code.

\newpage

\section{Implementation Details}
\label{app:implementation}

\paragraph{Search.}
We use a flat candidate pool with parent links and rank-normalized PUCT, with $c_{\mathrm{puct}}=0.25$ and a uniform prior. Each trajectory receives 100 authoring attempts.

\paragraph{Mechanism library.}
Mechanism extraction is triggered when a valid child improves over its parent by at least $\delta=0.02$. Up to the eight most recent library entries are included in each refinement prompt.

\paragraph{Foundation models.}
Canonical experiments use Gemini 3.1 Pro
(\texttt{gemini-3.1-pro-preview}) \citep{deepmind2026gemini31pro}.
The backbone robustness experiment uses DeepSeek-V4 Pro (\texttt{deepseek-v4-pro}) \citep{deepseek2026v4}, while keeping the authoring harness and evaluation protocol fixed.

\paragraph{Generation settings.}
Blueprint generation uses temperature 0.2; root-simulator generation and refinement use temperature 0.3. Generated simulators expose explicit random seeds so that their outputs can be reproduced.

\paragraph{Visual comparison.}
Each refinement call in the \framework{} configuration receives a real-versus-synthetic visualization containing three representative samples generated with a fixed simulator seed. The same visual feedback is provided to Text-Opt. Failure to render the visualization does not invalidate an authoring attempt.

\paragraph{Waveform contracts.}
Each candidate generates 100 outputs per evaluation. Audio simulators produce 10-s signals evaluated after resampling to 16\,kHz; PPG simulators produce 30-s signals at 125\,Hz; and ECG simulators produce 10-s two-lead signals at 500\,Hz. The output contract, including format, sampling rate, duration, sample count,
and channel structure, is fixed throughout authoring.

\paragraph{Budget accounting.}
Invalid or non-executable revisions consume an authoring attempt. API timeouts and rate-limit failures are retried and do not consume the budget.

\newpage
\section{Benchmark Data}\label{app:datasets}

\begin{table}[h]
\centering
\caption{\textbf{Reference data for the six-condition benchmark.}
Search recordings are available during authoring; held-out recordings are used only for evaluation.
Counts are recordings/windows, with unique participants in parentheses where available.}
\label{tab:data}

\small
\ra{1.02}
\setlength{\tabcolsep}{4pt}

\begin{tabular}{@{}lllcc@{}}
\toprule
Condition & Split & Dataset & Modality & $n$ \\
\midrule

\multirow{2}{*}{VSD}
& Search   & ZCHSound~\citep{jia2024zchsound} & Cardiac audio & 60 (60) \\
& Held-out & ZCHSound & Cardiac audio & 40 (40) \\

\midrule
\multirow{2}{*}{AS}
& Search   & BMD-HS~\citep{ali2026buet} & Cardiac audio & 37 (5) \\
& Held-out & BMD-HS & Cardiac audio & 23 (3) \\

\midrule
\multirow{2}{*}{COPD}
& Search   & ICBHI 2017~\citep{rocha2017respiratory} & Resp. audio & 65 (25) \\
& Held-out & KAUH + TR~\citep{fraiwan2021dataset,altan2017multimedia} & Resp. audio & 99 (45) \\

\midrule
\multirow{2}{*}{AF}
& Search   & MIMIC PERform AF~\citep{charlton2022detecting} & PPG & 96 (12) \\
& Held-out & Long-term AF PPG~\citep{bacevicius2024longterm} & PPG & 40 (8) \\

\midrule
\multirow{2}{*}{LQT}
& Search   & PTB-XL~\citep{wagner2020ptbxl} & ECG & 95 (86) \\
& Held-out & Chapman--Shaoxing--Ningbo~\citep{zheng2022ecgarrhythmia}
& ECG & 394 \\

\midrule
\multirow{2}{*}{WPW}
& Search   & PTB-XL~\citep{wagner2020ptbxl} & ECG & 60 (49) \\
& Held-out & Chapman--Shaoxing--Ningbo~\citep{zheng2022ecgarrhythmia}
& ECG & 72 \\

\bottomrule
\end{tabular}
\end{table}

\newpage


\section{Multi-seed results}
\label{app:multiseed}

Unless otherwise stated, all stochastic authoring experiments are run with
three random seeds. For each method--task pair in the main comparison, we rank
the three runs by their final best-so-far score $S_{\max}@100$ and use the
median-scoring run for main-text reporting. Figure~\ref{fig:baseline} shows the
complete trajectory of this representative run, while
Table~\ref{tab:main} reports its endpoint statistics. This appendix reports the
complete seed-level results underlying the main comparison, focused ablations,
sensitivity analyses, and held-out evaluation.


\subsection{Main authoring results across seeds}
\label{app:multiseed-main}

Table~\ref{tab:per_seed_final} reports the final $S_{\max}@100$ for all three
runs underlying the main benchmark. The representative results in
Table~\ref{tab:main} and Figure~\ref{fig:baseline} correspond to the run with
the median final score for each method--task pair.

\begin{table}[ht]
\centering
\caption{
\textbf{Seed-level main authoring results.}
Final best-so-far score $S_{\max}@100$ for all three random seeds.
For each method--task pair, the main paper reports the run with the median
final score; Figure~\ref{fig:baseline} shows the complete trajectory of the
corresponding representative run. The final column gives the sample standard
deviation across seeds. Root $\mathcal{S}^{(0)}$ and Independent Sampling are
fixed references reported in Table~\ref{tab:main}.
}
\label{tab:per_seed_final}

\small
\setlength{\tabcolsep}{4.5pt}
\renewcommand{\arraystretch}{1.06}

\begin{tabular*}{0.88\textwidth}{
@{\extracolsep{\fill}}
llcccc
@{}
}
\toprule
\textbf{Task}
& \textbf{Method}
& \textbf{Seed 1}
& \textbf{Seed 2}
& \textbf{Seed 3}
& \textbf{Std. dev.} \\
\midrule

\multirow{3}{*}{VSD}
& PUCT Score Search & .365 & .356 & .337 & .014 \\
& Text-Opt          & .416 & .512 & .346 & .083 \\
& \framework        & .587 & .448 & .611 & .088 \\

\midrule

\multirow{3}{*}{AS}
& PUCT Score Search & .254 & .247 & .227 & .014 \\
& Text-Opt          & .370 & .412 & .504 & .069 \\
& \framework        & .429 & .559 & .557 & .074 \\

\midrule

\multirow{3}{*}{COPD}
& PUCT Score Search & .577 & .455 & .398 & .091 \\
& Text-Opt          & .529 & .588 & .549 & .030 \\
& \framework        & .672 & .521 & .689 & .092 \\

\midrule

\multirow{3}{*}{AF}
& PUCT Score Search & .549 & .640 & .560 & .050 \\
& Text-Opt          & .771 & .806 & .801 & .019 \\
& \framework        & .748 & .738 & .738 & .006 \\

\midrule

\multirow{3}{*}{LQT}
& PUCT Score Search & .395 & .655 & .532 & .130 \\
& Text-Opt          & .554 & .501 & .582 & .041 \\
& \framework        & .541 & .568 & .594 & .027 \\

\midrule

\multirow{3}{*}{WPW}
& PUCT Score Search & .550 & .516 & .589 & .037 \\
& Text-Opt          & .364 & .548 & .526 & .100 \\
& \framework        & .581 & .543 & .653 & .056 \\

\bottomrule
\end{tabular*}
\end{table}

\newpage

\subsection{Focused ablations across seeds}
\label{app:multiseed-ablation}

Table~\ref{tab:ablation_all_seeds} expands the focused ablations from the main
paper to all three random seeds. We report both endpoint quality,
$S_{\max}@100$, and normalized $\mathrm{AUC}_{100}$ of the best-so-far
trajectory, which captures performance across the full authoring budget.

\begin{table}[ht]
\centering
\caption{
\textbf{Seed-level focused ablations.}
Each cell reports Seed 1 / Seed 2 / Seed 3.
\textbf{(a)} Final best-so-far score $S_{\max}@100$.
\textbf{(b)} Normalized $\mathrm{AUC}_{100}$ of the best-so-far trajectory.
LQT and WPW use the final ECGFounder representation-space evaluator.
}
\label{tab:ablation_all_seeds}

\small
\setlength{\tabcolsep}{2.2pt}
\renewcommand{\arraystretch}{1.08}


\textbf{(a) $S_{\max}@100$}\par\vspace{2pt}

\begin{tabular*}{\textwidth}{
@{\extracolsep{\fill}}
lcccccc
@{}
}
\toprule
\textbf{Configuration}
& \textbf{VSD}
& \textbf{AS}
& \textbf{COPD}
& \textbf{AF}
& \textbf{LQT}
& \textbf{WPW} \\
\midrule

\textbf{SimAuthor}
& .587/.448/.611
& .429/.559/.557
& .672/.521/.689
& .748/.738/.738
& .541/.568/.594
& .581/.543/.653 \\

$-$ Report
& .417/.366/.270
& .152/.452/.152
& .603/.609/.648
& .638/.548/.589
& .448/.460/.463
& .491/.497/.492 \\

$-$ Library
& .399/.409/.283
& .383/.447/.287
& .662/.704/.540
& .719/.725/.703
& .627/.540/.488
& .573/.555/.623 \\

\bottomrule
\end{tabular*}

\vspace{7pt}


\textbf{(b) Normalized $\mathrm{AUC}_{100}$}\par\vspace{2pt}

\begin{tabular*}{\textwidth}{
@{\extracolsep{\fill}}
lcccccc
@{}
}
\toprule
\textbf{Configuration}
& \textbf{VSD}
& \textbf{AS}
& \textbf{COPD}
& \textbf{AF}
& \textbf{LQT}
& \textbf{WPW} \\
\midrule

\textbf{SimAuthor}
& .359/.395/.497
& .415/.461/.458
& .544/.481/.599
& .699/.723/.717
& .521/.539/.549
& .574/.528/.614 \\

$-$ Report
& .356/.359/.269
& .143/.355/.141
& .456/.589/.585
& .574/.501/.556
& .439/.447/.431
& .465/.486/.492 \\

$-$ Library
& .348/.394/.282
& .327/.324/.281
& .528/.559/.529
& .668/.686/.687
& .576/.514/.474
& .547/.522/.599 \\

\bottomrule
\end{tabular*}

\end{table}


\subsection{Held-out evaluation across seeds}
\label{app:multiseed-heldout}

Held-out evaluation follows a different aggregation protocol because held-out
recordings must not influence model or seed selection. For each authoring seed,
we evaluate the simulator selected solely by its search score, without access
to the held-out set. Table~\ref{tab:heldout_seeds} therefore reports all three
independent runs and their column-wise median; no seed is selected using
held-out performance.

\begin{table}[ht]
  \centering
  \caption{
  \textbf{Seed-level held-out evaluation.}
  Each \framework{} row evaluates the search-score-best simulator from one
  independent authoring seed; held-out recordings are never used for node or
  seed selection. Root and best-of-100 sampling are fixed references.
  The final row reports the column-wise median across the three \framework{}
  runs and gives the values reported in Table~\ref{tab:heldout}.
  }
  \label{tab:heldout_seeds}

  \small
  \ra{1.05}
  \setlength{\tabcolsep}{4.2pt}

  \begin{tabular}{@{}lcccccc@{}}
  \toprule
  \textbf{Method}
  & \textbf{VSD}
  & \textbf{AS}
  & \textbf{COPD}
  & \textbf{AF}
  & \textbf{LQT}
  & \textbf{WPW} \\
  \midrule

  Root $\mathcal S^{(0)}$
  & .2229 & .1027 & .0875 & .1044 & .2771 & .3848 \\

  Sampling
  & .4246 & .1460 & .3624 & .1173 & .3408 & .4433 \\

  \midrule

  \framework{} seed 1
  & .5771 & .4264 & .6109 & .2040 & .4926 & .4965 \\

  \framework{} seed 2
  & .4093 & .4402 & .4095 & .2428 & .4245 & .5379 \\

  \framework{} seed 3
  & .5786 & .4737 & .5072 & .1813 & .4804 & .6155 \\

  \midrule

  \framework{} median
  & \textbf{.5771}
  & \textbf{.4402}
  & \textbf{.5072}
  & \textbf{.2040}
  & \textbf{.4804}
  & \textbf{.5379} \\

  \bottomrule
  \end{tabular}
\end{table}

\newpage

\section{Compute and token-matched controls}
\label{app:compute}

All methods in the main comparison receive the same budget of 100 authoring
attempts, but their model inputs differ substantially in length. In particular,
\framework{} carries the selected program, scientific blueprint, discrepancy
feedback, and mechanism memory across revisions. We therefore audit token usage
and additionally compare against independent sampling under a matched token
budget.

\subsection{Compute accounting}

Table~\ref{tab:compute} reports the model usage of a 100-attempt run, averaged
over conditions and completed seeds. Evaluation cost is identical per candidate
and is excluded. \framework{} uses approximately $3.8\times$ the total tokens
of independent sampling, almost entirely because of its larger input context;
output-token usage remains comparable. This corresponds to token parity with
best-of-100 sampling after roughly 26 \framework{} attempts, or to an
independent-sampling budget of approximately 380 proposals.

\begin{table}[ht]
\centering
\caption{
\textbf{Compute per 100-attempt run.}
Mean $\pm$ s.d.\ across conditions and completed seeds.
Token counts are estimated from character counts and exclude image tokens.
}
\label{tab:compute}

\small
\renewcommand{\arraystretch}{1.05}
\setlength{\tabcolsep}{5pt}

\begin{tabular}{@{}lcccc@{}}
\toprule
\textbf{Method}
& \textbf{Calls}
& \textbf{Input (k)}
& \textbf{Output (k)}
& \textbf{Total (k)} \\
\midrule

Independent Sampling
& 100.0
& $46 \pm 1$
& $185 \pm 22$
& 231 \\

PUCT Score Search
& 100.0
& $314 \pm 65$
& $270 \pm 68$
& 584 \\

\framework{}
& $106.4 \pm 1.4$
& $634 \pm 68$
& $244 \pm 71$
& 878 \\

\bottomrule
\end{tabular}
\end{table}

The additional \framework{} calls are auxiliary blueprint generation and
mechanism extraction after productive revisions. Because token counts are
estimated rather than taken from provider usage metadata, the absolute values
should be interpreted approximately; the large difference in input context is
the relevant comparison. Image inputs used by \framework{} are not counted, so
its reported input usage is a lower bound.

\subsection{Token-matched independent sampling}

To test whether the gain can be explained by the larger inference budget alone,
we expand independent sampling from 100 to 380 proposals per task, matching the
approximate token budget of a 100-attempt \framework{} run.

\begin{table}[ht]
\centering
\caption{
\textbf{Independent sampling at a token-matched budget.}
Best-of-$N$ over an approximately 380-program independent pool, compared with
\framework{} at its standard 100-attempt budget. ``Index of best'' gives the
position of the highest-scoring program within the sampling sequence.
}
\label{tab:sampling}

\small
\renewcommand{\arraystretch}{1.05}
\setlength{\tabcolsep}{4.5pt}

\begin{tabular}{@{}lcccccc@{}}
\toprule
& \textbf{VSD}
& \textbf{AS}
& \textbf{COPD}
& \textbf{AF}
& \textbf{LQT}
& \textbf{WPW} \\
\midrule

Best-of-100
& .477 & .184 & .391 & .492 & .418 & .456 \\

Best-of-380
& .479 & .298 & .475 & .530 & .505 & .458 \\

\quad index of best
& 118 & 371 & 254 & 324 & 313 & 338 \\

\midrule

\framework{}
& \textbf{.587}
& \textbf{.557}
& \textbf{.672}
& \textbf{.738}
& \textbf{.568}
& \textbf{.581} \\

\bottomrule
\end{tabular}
\end{table}

Increasing the independent pool from 100 to 380 improves the best score only
modestly on most tasks, while \framework{} remains higher on all six.
The largest sampling gain occurs on AS, whose best proposal appears late in the
pool, but a substantial gap to \framework{} remains. Thus, the advantage of
iterative authoring is not explained by the additional token budget alone.

\newpage 

\section{Downstream ECG augmentation}
\label{app:ecg-downstream}

We provide additional details for the downstream ECG experiment in
Section~\ref{sec:generalize}. The experiment tests whether ECGs produced by an
authored simulator improve classification of real recordings, rather than
merely matching the representation-space objective used during authoring.

\paragraph{Classification setup.}
We consider LQT and WPW as binary classification tasks. All conditions use the
same real PTB-XL training set, frozen ECGFounder representation, and
class-weighted logistic-regression probe. Evaluation is performed on a
patient-disjoint PTB-XL split (ID) and on Chapman--Shaoxing as an external OOD
dataset. Augmented conditions add exactly 300 positive training examples; all
other representation extraction and classifier-training settings are held
fixed.

\paragraph{Conventional waveform augmentation.}
As a non-generative augmentation baseline, we sample positive LQT or WPW
training recordings with replacement and generate 300 additional positives.
For each sampled recording, one of three waveform perturbations is selected
with equal probability:
\begin{align}
\text{Gaussian noise:}\qquad
x' &= x + \epsilon,
&
\epsilon &\sim
\mathcal{N}\!\left(0,(0.01\sigma_x)^2\right),
\\
\text{Baseline wander:}\qquad
x' &= x + 0.05\sigma_x\sin(2\pi f t),
&
f &\sim \mathcal{U}(0.1,0.5)\;\text{Hz},
\\
\text{Amplitude scaling:}\qquad
x' &= ax,
&
a &\sim \mathcal{U}(0.9,1.1),
\end{align}
where $\sigma_x$ is the standard deviation of the source waveform.
These operations are common ECG augmentation primitives and are contained in
the augmentation space studied by Raghu et al.~\citep{raghu2022data}.
Our baseline is a fixed, simplified subset of that augmentation family rather
than a reproduction of their learned TaskAug policy.

\paragraph{SimAuthor augmentation and selection.}
For the SimAuthor condition, the 300 additional positive examples are sampled
from an authored simulator rather than obtained by perturbing existing
recordings. Among candidates produced across the three independent authoring
seeds, the simulator is selected solely by its representation-space authoring
score. 

\paragraph{Evaluation metrics.}
The main paper reports AUPRC in Table~\ref{tab:generalization}. We additionally
report AUROC in Table~\ref{tab:ecg_downstream_auroc}. Both metrics are computed
from the same trained probes and evaluation sets. AUROC reproduces the AUPRC
pattern: SimAuthor gives the strongest result in all four ID/OOD comparisons,
with the largest improvements occurring on the external OOD evaluation.

\begin{table}[ht]
\centering
\caption{
\textbf{AUROC for downstream ECG classification.}
All conditions use frozen ECGFounder representations and the same
class-weighted logistic-regression probe. ``+ Conv.'' and ``+ SimAuthor''
augment the real training set with 300 positive examples. The SimAuthor
program is selected using only its representation-space authoring score across
three independent authoring runs; downstream performance is never used for
selection. ID denotes patient-disjoint PTB-XL and OOD denotes
Chapman--Shaoxing. Bold indicates the best result in each column.
}
\label{tab:ecg_downstream_auroc}

\small
\renewcommand{\arraystretch}{1.08}
\setlength{\tabcolsep}{5pt}

\begin{tabular}{@{}lcccc@{}}
\toprule
& \multicolumn{2}{c}{\textbf{LQT}}
& \multicolumn{2}{c}{\textbf{WPW}} \\
\cmidrule(lr){2-3}
\cmidrule(l){4-5}

\textbf{Training}
& \textbf{ID}
& \textbf{OOD}
& \textbf{ID}
& \textbf{OOD} \\
\midrule

Real
& $.965$
& $.797$
& $.951$
& $.926$ \\

+ Conv.
& $.968$
& $.821$
& $.953$
& $.926$ \\

\textbf{+ SimAuthor}
& $\mathbf{.974}$
& $\mathbf{.889}$
& $\mathbf{.973}$
& $\mathbf{.965}$ \\

\bottomrule
\end{tabular}
\end{table}

\newpage
\section{Revision Audit}
\label{app:audit}

Each 100-attempt \framework{} run produces a branching tree of simulator candidates. Our audit does not treat all attempts as revisions of the final
artifact. Instead, for each of the 18 runs, we trace the ancestry of the selected simulator back to the root and audit every valid, scored
parent--child transition on that lineage. These selected lineages contain 138
revisions in total. 

For each revision, added and removed lines are first aligned
deterministically. Edits that become identical after masking numeric literals are marked as scalar changes; a separate labelling model (DeepSeek-V4-Flash) then determines whether the overall behavioral change is pure parameter tuning or structural. All audited programs were authored by
Gemini 3.1 Pro. The structural-gain statistic in
Figure~\ref{fig:code-analysis} is computed directly from the signed parent--child score deltas after this partition.

Mechanism reuse is assessed separately by matching stored mechanism code against later programs and verifying that the matched code implements the same behavior. By this criterion, 69 of 86 extracted mechanisms are reused. All authored simulator scripts are released together with the search lineages and audit metadata, enabling future work to inspect revisions and alternative branches directly.

\newpage
\section{Prompts}\label{app:prompts}

Below are the complete prompts used by \framework. Four are issued during authoring, in the order they occur in
Algorithm~\ref{alg:simauthor}: the scientific blueprint (Prompt 1, Line 1), the root simulator (Prompt 2, Line 2), refinement (Prompt 3, Line 8), and mechanism extraction (Prompt 4, Line 13). No prompt was tuned per condition; the same four templates are used for all six tasks.

Braced tokens such as \slot{condition\_name} are runtime slots filled by the
harness. The refinement prompt is the one that carries persistence, and its
slots map one-to-one onto the context tuple assembled in Line 8 of
Algorithm~\ref{alg:simauthor}: \slot{blueprint} is $B$, \slot{parent\_code} is
$\mathrm{code}(n_p)$, \slot{feedback\_report} is $\mathrm{report}(n_p)$, and
\slot{mechanism\_context} is the most recent at most eight entries of $M$
(\appref{app:implementation}). The visual comparison $\mathrm{visual}(n_p)$ is
attached to the same call as an image rather than as text, so it does not
appear in the template.

The \slot{modality\_specification} slot is a fixed per-modality document stating the immutable output contract: file format, sampling rate, duration, sample count, channel or lead structure, and naming convention. It is identical across all conditions within a modality.\footnote{The four modality specifications (cardiac audio,
respiratory audio, ECG, PPG) are not reproduced here because they are
configuration rather than instruction: they restate the waveform contracts
already summarised in \appref{app:implementation} and are held fixed
throughout every run and every method. They are released with the code, together with the per-condition blueprints these prompts produce and the search trees with every program and discrepancy report.}

\subsection{Scientific Blueprint}
\label{app:blueprint}

\begin{tcolorbox}[colframe=black,
  title=\small{Prompt 1: Scientific blueprint generation. One call per task, issued before any recording is consulted.}
]
\setlist{leftmargin=1.3em,topsep=2pt,itemsep=0.5pt,parsep=0pt}

Produce a comprehensive Scientific Blueprint for \slot{condition\_name}
(\slot{signal\_modality} signal). This document will be the sole clinical
reference for a DSP team building a computational simulator. \\

\textbf{Modality Specification:} \slot{modality\_specification} \\

\textbf{Required Sections} \\

\textbf{1. Condition Overview}
\begin{itemize}
    \item Definition, pathophysiology, and clinical significance.
    \item Key subtypes or severity gradations (if applicable).
    \item Prevalence and typical patient demographics.
\end{itemize}

\textbf{2. Signal Manifestation}

Describe in quantitative detail how \slot{condition\_name} manifests in the
\slot{signal\_modality} domain. Follow the Modality Specification above for
which properties to characterise and at what level of detail.

\textbf{3. Severity and Variability}
\begin{itemize}
    \item How the signal characteristics change with condition severity or disease stage.
    \item Sources of natural variability (subject-to-subject, beat-to-beat, recording-condition, acquisition-related, or other).
\end{itemize}

\textbf{4. Key Discriminators}
\begin{itemize}
    \item The 3--5 most important features that distinguish this condition from similar conditions in the same modality.
    \item For each: typical range, how it differs from confounds.
    \item Include a differential diagnosis table comparing against 2--4 similar conditions.
\end{itemize}

\textbf{5. Synthesis Considerations}
\begin{itemize}
    \item Recommended computational approach for reproducing this condition in the \slot{signal\_modality} domain.
    \item Concrete parameter ranges the DSP engineer should target.
    \item Common pitfalls and how to avoid them.
\end{itemize}

\textbf{Output Format.} Write in clear, technical markdown. Use tables for
comparisons and reference values. Be SPECIFIC with numbers. If you are
uncertain about a value, give your best estimate and mark it with (est.).
This document should be 800--1500 words --- comprehensive but focused.

\end{tcolorbox}

\newpage

\subsection{Root Simulator}
\label{app:root}

\begin{tcolorbox}[colframe=black,
  title=\small{Prompt 2: Root simulator $\mathcal S^{(0)}$. One independent call; no discrepancy report exists yet.}
]
\setlist{leftmargin=1.3em,topsep=2pt,itemsep=0.5pt,parsep=0pt}

Construct a self-contained Python simulator for:
\begin{itemize}
    \item \textbf{Clinical condition:} \slot{condition\_name}
    \item \textbf{Signal modality:} \slot{signal\_modality}
\end{itemize}

Simulate how the target condition manifests specifically in the requested
signal modality. Use your biomedical and computational knowledge to identify
the principal physiological or signal-generating mechanisms and implement them
as an executable simulator. \\

The simulator should provide a simple but scientifically meaningful starting
point for later refinement. It should:
\begin{itemize}
    \item organize the synthesis into interpretable components;
    \item expose important physiological and simulation parameters explicitly;
    \item generate exactly \slot{num\_samples} diverse samples;
    \item represent plausible variation in severity, physiology, subjects, and acquisition where appropriate;
    \item avoid near-identical outputs that differ only through trivial noise;
    \item remain modular, robust, and easy to revise;
    \item use an explicit random seed for reproducibility.
\end{itemize}

\textbf{Modality Specification:} \slot{modality\_specification} \\

\textbf{Output.} Return only a complete executable Python script in one fenced
\texttt{python} code block. Do not include any text outside the code block.
The script must run without manual modification as:

\begin{verbatim}
python script_000.py
\end{verbatim}

\end{tcolorbox}

\newpage

\subsection{Refinement}
\label{app:refine}

\begin{tcolorbox}[colframe=black,
  title=\small{Prompt 3: Refinement. One call per authoring attempt; this is the call that carries persistence.}
]
\setlist{leftmargin=1.3em,topsep=2pt,itemsep=0.5pt,parsep=0pt}
\small

You are a senior DSP engineer. Your task is to improve a biomedical signal
simulator by applying one focused, hypothesis-driven refinement. \\

\textbf{Task Identity.} Clinical condition: \slot{condition\_name}. Signal
modality: \slot{signal\_modality}. \\

\textbf{Scientific Reference.} \slot{blueprint} \\

\textbf{Immutable Output Contract.} \slot{modality\_specification} \\

You may change simulator mechanisms and internal parameters, but you must NOT
change the output format, sampling rates, duration, sample count, channel or
lead structure, units, filename convention, or expected waveform shape defined
in the contract above. The number of generated recordings, output format,
sampling rate, duration, and output interface are controlled by the framework
and are not simulator design variables. Do not modify them during refinement. \\

\textbf{Current Simulator} (\texttt{script\_}\slot{parent\_idx}):
\begin{verbatim}
```python
{parent_code}
```
\end{verbatim}

\textbf{Numerical Discrepancy Report.} \slot{feedback\_report} \\

\textbf{Mechanism Library.} \slot{mechanism\_context} \\

\textbf{Refinement Policy}
\begin{enumerate}
    \item Study the Numerical Discrepancy Report carefully. Identify the most plausible causes of the reported discrepancies in the current simulator.
    \item Formulate ONE primary scientific hypothesis about what change would most effectively address the largest or most scientifically meaningful discrepancy.
    \item Implement a FOCUSED MUTATION that tests this hypothesis:
    \begin{itemize}
        \item Change only what is necessary to test the hypothesis.
        \item Preserve all unaffected working components exactly as they are.
        \item Do NOT rewrite the entire script unless the parent is fundamentally invalid (non-executable, wrong modality, violates the output contract).
        \item Each refinement should represent one main scientific change, not a collection of unrelated edits.
    \end{itemize}
    \item The final output must respect the Immutable Output Contract above.
\end{enumerate}

\textbf{Important: Role of the Numerical Discrepancy Report.} The Search
Reference Set contains a limited sample of real recordings. It provides
empirical grounding but does not exhaustively define the target distribution.
Use the discrepancy report to identify clear unrealistic behaviours (e.g.,
missing physiological features, implausible morphologies, severe mode
collapse). However, scientifically plausible variability described in the
Scientific Blueprint should be preserved even if it is not fully represented
in the finite reference sample. Do not simply imitate the reference set
statistics. Address genuine discrepancies while respecting the Scientific
Blueprint's broader description of the condition. \\

\textbf{Output Format.} Return ONLY the complete, self-contained Python script
in one fenced \texttt{python} code block. Do not include explanations,
commentary, or Markdown outside the code block.

\end{tcolorbox}

\newpage

\subsection{Mechanism Extraction}
\label{app:extraction}

\begin{tcolorbox}[colframe=black,
  title=\small{Prompt 4: Mechanism extraction. One call per accepted improvement, i.e.\ when child $-$ parent $\geq \delta$.}
]
\setlist{leftmargin=1.3em,topsep=2pt,itemsep=0.5pt,parsep=0pt}

Compare two versions of a biomedical signal synthesis script. The child
achieved a better evaluation than the parent. \\

Identify ONE plausible reusable behavioral change associated with the
improvement. \\

Focus on substantive changes to signal generation, physiological modeling,
parameterization, temporal structure, spectral structure, waveform morphology,
or stochastic variation. \\

Ignore changes limited to formatting, comments, logging, file paths, code
organisation without behavioral effect, or error handling. \\

Use the unified diff below as the primary evidence:
\begin{verbatim}
```diff
{diff}
```
\end{verbatim}

Return ONLY a single-line JSON object (no Markdown, no literal newlines inside
string values):
\begin{verbatim}
{"name": "snake_case_3_5_words",
 "mechanism": "one-sentence description",
 "code_snippet": "key function or short description"}
\end{verbatim}

\end{tcolorbox}

\end{document}